%% file: main.tex
\RequirePackage[2020-02-02]{latexrelease}
\documentclass{clv3}

\usepackage{hyperref}
\usepackage[table]{xcolor}
\definecolor{darkblue}{rgb}{0, 0, 0.5}
\hypersetup{colorlinks=true,citecolor=darkblue, linkcolor=darkblue, urlcolor=darkblue}

\usepackage{graphicx}
\usepackage{caption}

\usepackage{amsmath, amsfonts}
\usepackage{varwidth}
\usepackage{capt-of}
\usepackage{booktabs}
\newsavebox\tmpbox
\usepackage{caption}
\usepackage{subcaption}

\newcommand{\hlc}[2][yellow]{{%
    \colorlet{foo}{#1}%
    \sethlcolor{foo}\hl{#2}}%
}

\DeclareMathOperator*{\argmax}{argmax}

\usepackage{makecell}
\usepackage{tabularx}
\usepackage{lipsum}
\usepackage{mwe}

\usepackage{algorithm} 
\usepackage{algpseudocode}

\usepackage{microtype}

\usepackage{booktabs}
\usepackage{listings}
\usepackage{xstring}
\usepackage{soul}
\lstdefinestyle{odin-style}{%
  language=odin,
  basicstyle=\ttfamily,
  xleftmargin=0pt,
  xrightmargin=0pt,
  frame=tb,
  rulecolor=\color{black},
  extendedchars=true,
  showstringspaces=false,
  showspaces=false,
  tabsize=2,
  breaklines=true,
  showtabs=false,
  escapeinside={(*@}{@*)}
}

\lstnewenvironment{odinrule}[1]{%
 \lstset{style=odin-style}}{%
\captionof{lstlisting}{#1}
}

\lstdefinelanguage{odin}{
    morestring=[b]{"},
    moredelim=*[s][]{\[}{\]},
    morestring=*[d]{=},
    morestring=*[d]{^},
    keywords=[1]{true, false, null, name, action, example,
    priority, type, label, unit, pattern}, 
    keywordstyle=[1]\bfseries,
    alsoletter={^><=?*}, 
    keywords=[3]{lemma, word, chunk, tag},
    keywordstyle=[3]\color{purple}\textbf,
    keywords=[4]{trigger, ?<trigger>,
        subject, object},
    keywordstyle=[4]\color{blue}\textbf, 
    keywords=[5]{%
        Protein,
    },
    keywordstyle=[5]\color{black}\textbf,
    comment=[l]{\#},
    commentstyle=\color{coolgrey}\textit,
    stringstyle=\ttfamily,
  sensitive=true 
  }
\algrenewcommand\algorithmicrequire{\textbf{Input:}}
\algrenewcommand\algorithmicensure{\textbf{Output:}}

\begin{document}
\issue{1}{1}{2016}

\runningtitle{Multitask Learning of Neural Relation and Explanation Classifiers}

\runningauthor{Tang et al.}

\pageonefooter{Action editor: Vivek Srikumar.
Submission received: 30 March 2022;
Revised version received: 31 August 2022;
Accepted for publication: 9 September 2022.}

\title{It Takes Two Flints to Make a Fire: \\Multitask Learning of Neural Relation and Explanation Classifiers}

\author{Zheng Tang\thanks{E-mail: zhengtang@arizona.edu.}}
\affil{University of Arizona}

\author{Mihai Surdeanu}
\affil{University of Arizona}

\maketitle

\begin{abstract}
We propose an explainable  approach for relation extraction that mitigates the tension between generalization and explainability by jointly training for the two goals. Our approach uses a multi-task learning architecture, which jointly trains a classifier for relation extraction, and a sequence model that labels words in the context of the relation that explain the decisions of the relation classifier. We also convert the model outputs to rules to bring global explanations to this approach.
This sequence model is trained using a hybrid strategy: supervised, when supervision from pre-existing patterns is available, and semi-supervised otherwise. In the latter situation, we treat the sequence model's labels as latent variables, and learn the best assignment that maximizes the performance of the relation classifier.
We evaluate the proposed approach on the two datasets and show that the sequence model provides labels that serve as accurate explanations for the relation classifier's decisions, and, importantly, that the joint training generally improves the performance of the relation classifier.
We also evaluate the performance of the generated rules and show that the new rules are great add-on to the manual rules and bring the rule-based system much closer to the neural models.
\end{abstract}

\input{sections/introduction}

\input{sections/relatedwork}

\input{sections/approach}

\input{sections/experiments}
\input{sections/conclusion}
\appendix
\input{sections/appendix}
\begin{acknowledgments}
We thank the reviewers and action editor for their thoughtful comments and suggestions.
This work was partially supported by the Defense Advanced Research Projects Agency (DARPA) under the World Modelers program, grant \#W911NF1810014, and by the National Science Foundation (NSF) under grant \#2006583. 
Mihai Surdeanu declares a financial interest in \url{lum.ai}. This interest has been properly disclosed to the University of Arizona Institutional Review Committee and is managed in accordance with its conflict of interest policies.
\end{acknowledgments}
\bibliography{refs}
\end{document}

%% file: sections/introduction.tex
\section{Introduction}


Many domains such as medical, legal, or finance, require that decision making be not only accurate but also trustworthy. Thus,  understanding what the underlying model captures is a critical requirement in such applications. 
To this end, previous efforts addressed this limitation by adding explainability to neural models, which have come to dominate natural language processing (NLP) \cite{manning-2015-last}. These explanations can be categorized along two main aspects: whether they explain a complete model ({\em global}) or individual predictions ({\em local}); and whether they are an integral part of the classification model itself ({\em self-explaining}) or are generated through a post-processing step ({\em post-hoc}) (see Related Work for a longer discussion).
Most of recent proposed efforts focus on the {\em local} and {\em post-hoc} explanations \cite{ribeiro2016should, P-295, schwab2019cxplain}. These directions have a few advantages such as modularity and simplicity. 
However, they also have two important drawbacks: these types of explanations are not guaranteed to be {\em faithful} to the original model to be explained, and are not {\em actionable}, i.e., even if they correctly explain an imperfect classification, there is no clear path towards correcting the underlying model because ``changing one thing changes everything'' in a neural network~\cite{sculley2015hidden}.


Our article focuses on addressing the limitations of these local and post-hoc explainability approaches by providing a self-explanatory neural architecture (i.e., explanations are part of classification) that can provide both local and global explanations.
In particular, we propose an approach for relation extraction (RE) that {\em jointly} learns how to explain and predict. Intuitively, our approach trains two classifiers: an explainability classifier (EC), which labels words in the textual context where the relation is expressed as important or not for the relation to be extracted, and a relation classifier (RC), which predicts the relation that holds between two given entities using {\em only} the words deemed as important. As such, our approach is {\em self-explanatory} because of inter-dependency between RC and EC, and generates {\em faithful} explanations that correctly depict how the relation classifier makes a decision \cite{vafa2021rationales}. 

The contributions of this article are the following:
{\flushleft {\bf (1)}} 
We introduce a hybrid strategy to jointly train the EC and RC. Our method trains the EC as a supervised classifier when information about which words are important for a relation exists. For example, in this article we use a small set of linguistic rules to identify the important words in the relation's context. For example, in the sentence {\em ``John was born in France,''} such a rule may identify the words {\em born} and {\em in} as important. 
Importantly, our approach requires minimal supervision for explanations, e.g., we report results when using an average of 7 rules per relation type on one dataset and fewer on another dataset. 
For the more common situation where training examples are not associated with such rules, we train using a semi-supervised strategy: we treat EC's labels as latent variables, and learn the best assignment that maximizes the performance of the RC. 

{\flushleft {\bf (2)}} 
We evaluate our approach on two datasets: TACRED~\cite{zhang2017tacred} and CoNLL04~\cite{roth-yih-2004-linear}. For (partial) explainability information, we  select from the surface rules provided with the dataset~\cite{zhang2017tacred,chang2014tokensregex} as well as from a small set of syntactic rules developed in-house using the Odin framework~\cite{valenzuela-escarcega-etal-2016-odins}. 
Our evaluation demonstrates that jointly training for prediction and explainability improves the performance of the relation classifier considerably on CoNLL04, and maintains the same level of performance on TACRED when compared with a state-of-the-art neural relation classifier. 
Importantly, our method achieves its best performance when using an average of 7 rules per relation type on TACRED and 4 rules per relation type for CoNLL04, which indicates that only minimal guidance from such rules is needed.
{\flushleft {\bf (3)}}
More relevant for the goals of this work, we also evaluate our method for explainability using two strategies. The first strategy is automated and focuses on the capacity of our method to identify the same words in the context as the ones identified by rules, to verify that our approach indeed encodes the proper linguistic knowledge. Thus, this evaluation looks at examples associated with rules. In this situation, we measure the overlap between the words identified by the EC as important and the words used by rules using standard precision, recall, and F1 scores. The second strategy relies on {\em plausability}, i.e., can the machine explanations be understood and interpreted by humans \cite{wiegreffe-pinter-2019-attention,vafa2021rationales}? To this end, we compare the tokens identified by the EC against human annotations of the context words marked as important for the relation. In both evaluations, our approach achieves considerably higher overlap with rules/human annotations than other strong baselines such as saliency mapping\mbox{~\cite{simonyan2013deep}}, LIME~\cite{ribeiro2016should}, SHAP~\cite{NIPS2017_7062}, CXPlain~\cite{schwab2019cxplain},  and greedy rationales~\cite{vafa2021rationales}.
{\flushleft {\bf (4)}}
We also explore the feasibility of transforming the local explanations into global ones. That is, instead of using the EC to explain individual predictions, we introduce a simple algorithm that converts the tokens marked as important into a set of rules that becomes a new, fully-explainable model that approximates the behavior of the neural RC. 
We compare the performance of this rule-based model with the performance of the rules written by domain experts, as well as with the neural RC model. The results show that our rule-based model has a considerably higher performance that the manually-written rules, approaching the performance of the neural classifier within a reasonable gap. In some real-world scenarios, this gap may be an acceptable cost, as the generated rule-based model provides {\em actionable} explainability. That is, when a rule is incorrect, a domain expert can improve it without impacting other parts of the models \cite{valenzuela2016snaptogrid}.

%% file: sections/relatedwork.tex
\section{Related Work}
Our work lies at the intersection of relation extraction and explainability. 
We summarize these two research areas next.

\subsection{Relation Extraction} 
Information extraction (IE), i.e., extracting structured information from text such as events and their participants, is one of the fundamental tasks in NLP that was shown to be useful for many end-user applications such as question answering \cite{srihari1999information, srihari2000question}, and summarization \cite{rau1989information, zechner1997literature}.
Our work focuses on a subtask of IE: relation extraction (RE), which addresses the extraction of (mostly) binary relations between entities such as {\tt place\_of\_birth}, which connects a person named entity with a location.

RE has received tremendous attention in the past several decades. We group the works on RE into two categories: before the ``deep learning tsunami'' \cite{manning-2015-last}, and after. 

\subsubsection{Relation extraction before deep learning}
 
 The first approaches for RE were rule-based. For example, \citet{hearst-1992-automatic} proposed a method to learn hyponymy relations using hand-written patterns. 
 \citet{riloff1996automatically} introduced a pattern acquisition method that alternates between learning paterns and extracting relation mentions. 
 \citet{Brin1998ExtractingPA} proposed a dual iterative pattern/relation expansion, which exploited the duality between patterns and relations. \citet{Hassan2006UnsupervisedIE} used Hyperlink-Induced Topic Search (HITS) \cite{Kleinberg1999HubsAA} to jointly learn patterns and relations in an unsupervised manner. In general, these rule-based methods usually obtain high precision but suffer from low recall. 
 While our explanations can be interpreted as rules, our work differs from these directions in two significant ways. First, most of these directions are iterative, alternating between learning patterns (or rules) and relations. In contrast, our approach trains relation and explanation classifiers jointly. Second, and probably more importantly, we show that our explanations often focus on parts of speech that are necessary for plausability (according to the human annotators) but are semantically-ambiguous such as prepositions and determiners. On the other hand, most pattern acquisition methods usually focus on clear syntactic structures such as subject-verb-object and words with more clear semantics such as nominals and verbs.

Statistical methods that followed the above rule-based approaches address the limited generality of rules. 
In terms of supervision, ``traditional'' machine learning approaches for RE include fully supervised methods~\cite{zelenko2003kernel, bunescu2005shortest}, or methods that rely on distant supervision, where training data is generated automatically by (noisily) aligning existing knowledge bases with texts~\cite{mintz2009distant,riedel2010modeling,hoffmann2011knowledge,surdeanu-etal-2012-multi}.
Most of these approaches used explicit features such as lexical, syntactic, and semantic. For example, \citet{kambhatla2004combining} proposed a maximum entropy classifier using these features. \citet{zhou-etal-2005-exploring} found that additional features such as syntactic chunks further help the classification performance. 
{\citet{jiang-zhai-2007-systematic}  evaluate the effectiveness of different feature spaces for RE. Similarly, \citet{chan-roth-2011-exploiting} expanded feature representations to include syntactico-semantic structures that improve RE.

Our work is conceptually similar to the method of \citet{chan-roth-2011-exploiting}. Similar to them, we extract relations only from the smaller context identified by a distinct component (the explainability classifier in our case). However, there are several important differences between these two efforts. First, the method of \citet{chan-roth-2011-exploiting} operates as a pipeline: they start by matching syntactico-semantic structures potentially indicative of relations, and then they apply a relation classifier only on the texts that match them. In contrast, our method jointly trains the relation and explainability classifiers. Second, the syntactico-semantic structures in \cite{chan-roth-2011-exploiting} were manually extracted and categorized, whereas our explanations are learned in a semi-supervised way from data and a small number of rules. Last but not least, the patterns of \citet{chan-roth-2011-exploiting} are non-lexicalized. In contrast, the explanations produced by our explainability classifier are lexicalized, which is critical for human understanding. 

Kernel methods were also a popular direction for relation extraction due to their advantage of avoiding feature engineering. 
To this end, \citet{miller-etal-2000-novel} introduced a sequence kernel for relation extraction. 
Several researchers proposed kernels designed around constituent parse trees to capture sentence grammatical structure \cite{miller-etal-2000-novel, zelenko2003kernel, moschitti2006making}. 
\citet{bunescu2005shortest, nguyen-etal-2009-convolution} introduced kernels based on syntactic dependencies, a simpler representation that flattens constituent trees while preserving most syntactic information.
To combine the information captured by individual kernels that model different representations, \citet{zhao-grishman-2005-extracting} presented a composite kernel which combines multiple such individual kernels.

\subsubsection{Deep learning methods for relation extraction}

Deep learning approaches for RE that rely on sequence models range from using CNNs or RNNs~\cite{zeng2014relation, zhang2015relation}, to augmenting RNNs with different components~\cite{xu-etal-2015-classifying, zhou-etal-2016-attention-based}, or to combining RNNs and CNNs~\cite{vu-etal-2016-combining, wang-etal-2016-relation}. Other approaches take advantage of graph neural networks \cite{zhang-etal-2018-graph} or attention mechanisms~\cite{zhang2017tacred}.

More recently, transformer-based~\cite{vaswani2017attention} approaches have shown considerable improvements on many natural language tasks including RE. For example, \citet{wu2019enriching} applied BERT~\cite{devlin2018bert} to the TACRED RE task. ~\citet{devlin2018bert, yamada2020luke} showed that further improvements are possible with a better representation for the pre-trained language model.

Our approach also fits in this space. We deploy a transformer-based classifier to capture relation mentions, but we also include a novel component dedicated to explainability, which tags the words important for the relation at hand. Importantly, our direction has the relation classifier operate directly on top of the words deemed important for the relation by the explainability classifier, which guarantees that our explanations are {\em faithful}, i.e., our explanations correctly depict how the relation classifier makes a decision \cite{vafa2021rationales}. Further, we propose an efficient semi-supervised strategy to jointly train the relation and explainability classifiers using a small amount of linguistic supervision for explainability.

\subsection{Explainability} 

Explainable artificial intelligence (XAI) has recently experienced a resurgence in the context of deep learning~\cite{adadi2018peeking, gunning2019darpa, arrieta2020explainable, danilevsky-etal-2020-survey}.

\subsubsection{A taxonomy of explanations}
Explanations can be categorized along two main aspects: whether they explain a complete model ({\em global}) or individual predictions ({\em local}); and whether they are built in the classification model itself ({\em self-explaining}) or are generated through a post-processing step ({\em post-hoc}).

\paragraph{Global vs. Local}
Rule-based approaches \cite{hearst-1992-automatic, Brin1998ExtractingPA} or decision trees \cite{bechet2000tagging, boros-etal-2017-fast} provide global explainability by constructing transparent models that people can understand.  
However, these directions were slowly replaced by deep learning, which tends to yield better classifiers (at least with respect to accuracy). Several efforts aimed at bringing back global explainability into deep learning.
For example, in the non-NLP context of high-stakes decision-making at population level, \citet{rawal2020beyond} proposed a model-agnostic framework that constructs global counterfactual explanations that provide an interpretable and accurate summary of recourses for an entire population affected by a certain problem such as bad financial credit.
Closer to our work, \mbox{\citet{craven1996extracting,frosst2017distilling}} both proposed distilling a neural network into a globally-interpretable model such as a decision tree. 

However, most recent approaches focus on local model explainability, which preserves the underlying neural classifier and interprets its individual predictions. In this category, \citet{hendricks2016generating} produced natural language explanations of individual model outputs. \citet{han-etal-2020-explaining} used influence-based training-point ranking to study spurious training artifacts in NLP settings. \citet{wachter2018counterfactual, karimi2020modelagnostic} used counterfactual explanations to understand model decisions. 

\paragraph{Self-explaining vs. Post-hoc}
Self-explaining strategies make explanations an integral part of model predictions. For example, \citet{tang-etal-2020-exploring} proposed an encoder-decoder method for relation extraction, which jointly classifies relations and decodes rules that explain the relation classifier's decisions. \citet{rajani-etal-2019-explain} proposed a framework that provide both answer and explanation for a commonsense QA task.
In contrast, post-hoc explanations include an additional component that generates explanations after the main model produces its decisions. In this space, \citet{10.1145/3219819.3220001} learn a taxonomy post-hoc to better interpret network embeddings. 
As mentioned above, \citet{craven1996extracting,frosst2017distilling} both proposed post-hoc strategies to distill neural network into decision trees. 
\citet{li-etal-2016-visualizing, fong2019understanding, hoover-etal-2020-exbert} provided post-hoc visualizations as model explanations.  \citet{belinkov-etal-2017-neural, peters-etal-2019-tune, zhao-bethard-2020-berts, hewitt-etal-2021-conditional} introduced probes, i.e., models trained to predict certain linguistic properties in order to verify that the underlying neural models have learned the desired linguistic knowledge.

With respect to this taxonomy, our approach is self-explaining because our relation extractor has access solely to the context identified as important by the explainability classifier, and local because our core method explains individual predictions. However, in the latter part of this article we propose a simple strategy that converts local explainability into global by converting the entire neural model into a set of rules using the words deemed as important in a dataset by the explainability classifier. 

\subsubsection{Finding rationales}
From a different perspective, our approach can be seen as finding {\em rationales}, i.e., subsets of context that explain individual model decisions \cite{vafa2021rationales}. Although these directions fit under local explainability (and mostly post-hoc) we discuss them separately due to their recent popularity and proximity to our work. 

Some efforts in this space used gradient-based saliency mapping to determine the importance of tokens in context \cite{baehrens2010explain, simonyan2013deep, devlin2018bert, voita2021analyzing}. However, gradients can be saturated, i.e., they may be close to zeros and, thus, lose explanatory signal. \citet{ghorbani2019interpretation, wang2020gradient} also warn that gradients are fragile and they can be distorted while keeping the same prediction.  

As an alternative, some researchers focused instead on attention weights in transformer networks \cite{wiegreffe2019attention, mohankumar-etal-2020-towards}. 
However, there is also evidence that
attention weights may not be good explanations~\cite{jain2019attention, brunner2019identifiability, kobayashi-etal-2020-attention}. Other efforts have used adversarial attacks on inputs to identify their importance. For example, HotFlip~\cite{ebrahimi2017hotflip} used word-level substitutions to impact predictions. CXPlain~\cite{schwab2019cxplain} calculates feature importance by masking them and comparing differences in output confidences.~\citet{feng2018pathologies, Li2016UnderstandingNN} focused on input reduction to identify the importance of input features. Instead of reducing, ~\citet{vafa2021rationales} greedily added input information to locate meaningful rationales. However, other research has showed that input perturbation cannot always guarantee a good explanation~\cite{poerner2018evaluating}. 

In a different direction, surrogate approaches~\cite{ribeiro2016should, NIPS2017_7062} generated artificial data in the neighborhood of a prediction to be explained, by randomly hiding features from the instance and learning a surrogate model to explain the predictions. AllenNLP~\cite{wallace2019allennlp} combined adversarial attacks and gradient-based saliency mapping in their toolkit. Lastly, \citet{lei-etal-2016-rationalizing, situ-etal-2021-learning} trained a generator model to produce feature importance. 

Other than the problems we mentioned above, most of these approaches are either passively reflecting the model behavior or learning rationales in an unsupervised way. Because of this, these methods cannot guarantee faithfulness and plausibility. In contrast, our proposed approach provides local explanations (or rationales) that are designed to be faithful. Further,  our empirical evaluation shows that our explanations are also more plausible than other rationale finding methods (see Section~\ref{sec:experiments}).

All of the approaches discussed above address the task of finding rationales. However, a relatively new direction focuses on the opposite effort: if rationales are provided by a human expert, how can they be integrated in a statistical model? For example,
~\citet{bao2018deriving} proposed a method to map discrete rationales to continuous attention, and showed that the performance on low-resource tasks can be improved by transferring these mappings from resource-rich tasks.
~\citet{hancock2018training} showed that human-provided natural language explanations for labeling decisions can be converted to noisy labels using a semantic parser. They empirically demonstrated that through this process they can train classifiers with comparable F1 scores considerably faster.
Incorporating rationales in a classifier is a key part of the our approach. However, our method jointly trains the explanation classifier with the relation classifier, rather than depending on human rationales for the entire training data.

%% file: sections/approach.tex
\section{Approach}
\label{approach}
At a high level, our approach consists of two main components: a neural relation classifier with an integrated explainability classifier, and a rule generation component, which generates a rule-based model from the explainability information, i.e., context words that explain a relation, provided by the neural model. 

\subsection{Walkthrough Example}
Before getting into the details of our approach, we highlight its key functionality with the walkthrough example shown in Table~\ref{tab:walk}.
\begin{table*}[]
\centering
\begin{tabular}{l}
\hline
  \raisebox{-\totalheight}{\includegraphics[width=0.9\textwidth]{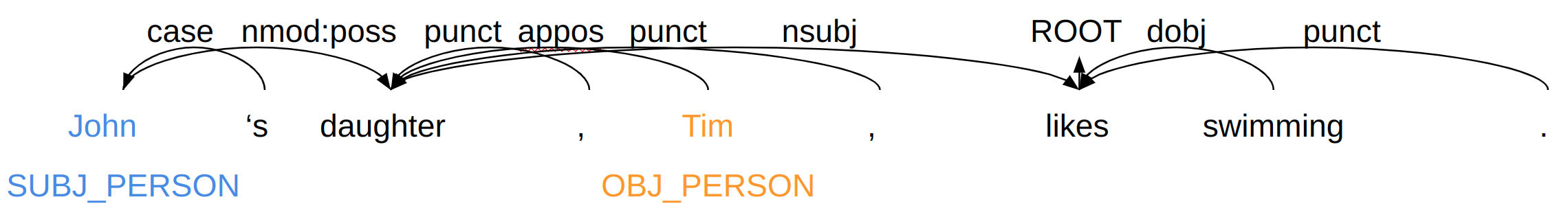}}\\
 \hline
 \multicolumn{1}{c}{(a) Input} \vspace{3mm}\\
 \hline
 \raisebox{-\totalheight}{\includegraphics[width=0.9\textwidth]{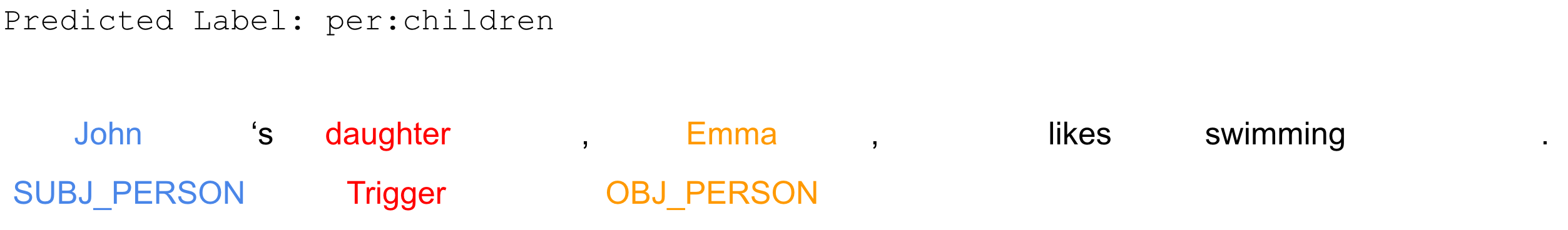}}\\
 \hline
 \multicolumn{1}{c}{(b) Relation classification and explainability outputs} \vspace{3mm}\\
 \hline
 \raisebox{-\totalheight}{\includegraphics[width=0.9\textwidth]{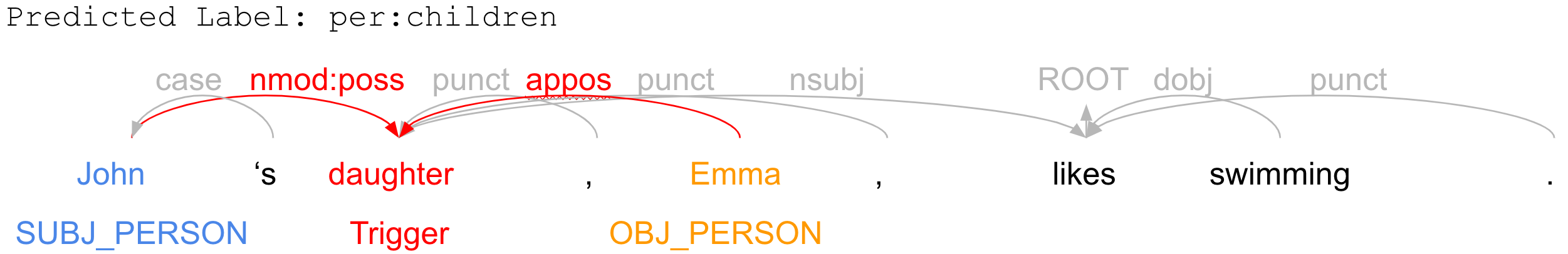}}\\
  \hline
 \multicolumn{1}{c}{(c) Information gathering for rule generation} \vspace{3mm}\\
 \hline
{\begin{lstlisting}[basicstyle=\ttfamily, escapechar=;]
;\colorbox{yellow!12}{\textbf{label}: per:children};
pattern: |
 trigger = 
    ;\colorbox{blue!12}{[word=/daughter/]};
 ;\colorbox{red!12}{$\,$\textbf{\textcolor{blue}{subject}}:$\,$SUBJ\_Person}; = nmod:poss
 ;\colorbox{red!12}{$\,$\textbf{\textcolor{blue}{object}}:$\,$OBJ\_Person}; = appos
\end{lstlisting}}
\\
 \hline
\multicolumn{1}{c} {(d) Generated rule} \vspace{3mm}\\
 \hline
\end{tabular}
\caption{Walkthrough example of our approach. 
The task input includes information about the entities participating in the relation (denoted as subject and object) and their types ({\tt PERSON} here). 
Our neural architecture, which includes both a relation and explanation classifier, predicts the relation that holds between the two entities ({\tt per:children} here, i.e., the object is the child of the subject), as well as which words best explain the decision (in red). 
In step (c), the rule generator collects the necessary information from the annotated sentence, i.e., the shortest syntactic dependency path that connects the two entities with the explanation words (in red in the figure). Step (d) shows the generated rule in the Odin language.
}
\label{tab:walk}
\end{table*}
Consider the sentence ``{\em John's daughter, Emma, likes swimming.}''.
As shown in Table \ref{tab:walk} (a), the task input includes: the raw text in the sentence, the entities participating in the relation (denoted as subject and object) and their types ({\tt PERSON} here), and the syntactic dependency parse tree.
Table \ref{tab:walk} (b) shows the output of our relation classifier (RC) and explanation classifier (EC):  the RC returns the predicted relation {\tt per:children}, while the EC labels the word {\em daughter} as the trigger of the predicted relation.
Step (c) shows the information that is collected for rule generation. This information includes: the two entities, the relation predicted, the tokens identified by the EC as the rationale for the relation, and the shortest syntactic path connecting the two entities with the rationale words. The output rule generated by our approach is shown in step (d). 
This rule is written in the Odin language \cite{Valenzuela:15,valenzuela-escarcega-etal-2016-odins}. The rule captures the relation to be predicted ({\tt per:children}), its trigger ({\em daughter}), the two arguments and their type, e.g., {\tt subject} with the type {\tt SUBJ\_Person}, and the syntactic paths between each argument and the trigger phrase, e.g., {\tt nmod:poss} for the subject argument. Note that in this simple example, the trigger consists of a single word, but, in general, an Odin rule can take any arbitrary sequence of words as its trigger.

This example shows that our method can be deployed in two ways. First, one can use the joint RC and EC neural classifiers, which predict relations that hold between pairs of entities, as well as local explanations (or rationales) that explain the prediction. 
Alternatively, a different class of users may use the output of step (d), which, once applied on large text collections, contains a set of rules that describes multiple relation classes. This usage may be preferred in real-world situations that have to mitigate the ``technical debt'' of neural methods, i.e., reduce the cost of maintaining these models over time \cite{sculley2015hidden}. Although not within the scope of this work, other works have shown that rule-based methods for IE can be improved and maintained at a low cost \cite{valenzuela2016snaptogrid}. 

\subsection{Joint Relation and Explainability Classifiers}

\begin{figure}[t]
    \centering
    \includegraphics[width=0.70\textwidth]{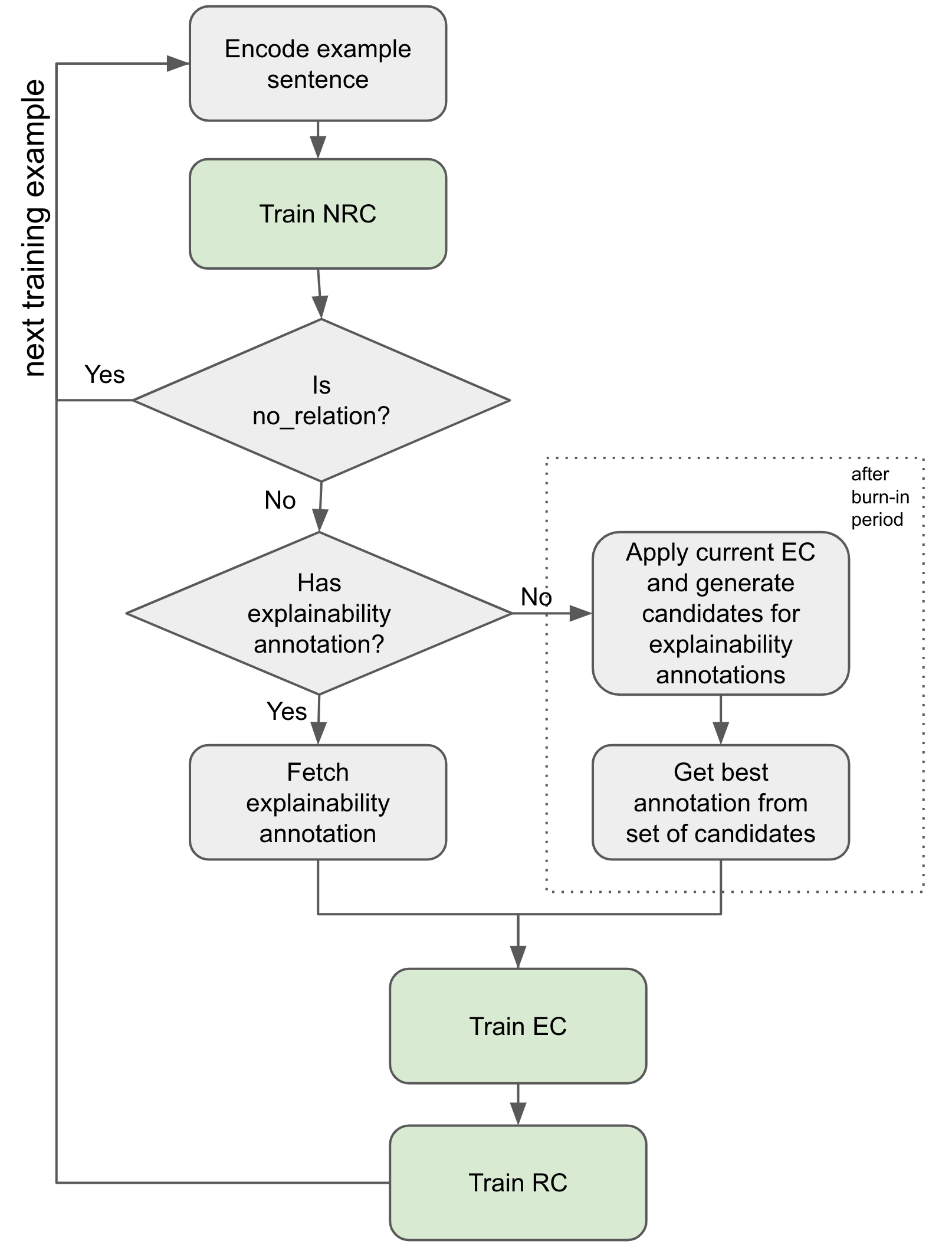}
    \caption{Flow of our semi-supervised training procedure for an individual training example. All the ``Train \dots'' blocks (green background) involve parameter updates of the corresponding classifiers. These updates are shown here for an individual training example, but are batched in the actual implementation.
    }
    \label{fig:flow}
\end{figure}

As mentioned, our approach jointly trains an explainability classifier (EC) and a relation classifier (RC). 
The RC is a multiclass classifier that distinguishes between actual relation labels seen in training. We couple the RC with a binary classifier that first predicts if the current example contains an actual relation or no relation (marked as {\tt no\_relation}). For conciseness, we call this classifier the no relation classifier (NRC). 
The EC is a binary word-level classifier, which labels words in the sentence that contains the relation with 1, if they are important for the underlying relation, or 0, otherwise. 

We start this section with the description of the overall training procedure, and follow with details about the individual classifiers.

\subsubsection{Training Procedure}
\label{sec:training}
The overall flow of the training procedure is shown in Figure~\ref{fig:flow}. This flow is temporally split in two periods: a burn-in period, which is fully supervised, followed by a period that includes semi-supervised learning (SSL).
This distinction is necessary because while all training examples in this task are guaranteed to have RC labels, most examples will {\em not} have gold explainability annotations. For example, for the sentence {\em ``[CLS] John was born in London.''}, the training data contains information that there is a {\tt per:city\_of\_birth} relation between {\em John} and {\em London}, but may not contain information about which words are critical for this relation ({\em born} and {\em in}). 



\paragraph{Burn-in period} In this stage, shown in the left-hand side of Figure~\ref{fig:flow}, we only use the training examples that are associated with explainability annotations (see Section~\ref{sec:expann} for details on how these annotations are generated). Here we train initial versions of the three classifiers: NRC, EC, and RC (see Section~\ref{sec:models} for details on the three classifiers). 
The purpose of this stage is to initialize the three classifiers such that they can be successfully used to reduce the search space for explainability annotations in the next SSL stage. 

\paragraph{After burn-in} In this stage, the training procedure is exposed to all training examples, including those without annotations for explainability. That is, for such training examples, we simply have annotations for the relation labels (or {\tt no\_relation}), without knowing which context words explain the underlying relation. In such situations, the right-hand side of the flow in Figure~\ref{fig:flow} is used, which triggers two additional components: one to generate candidates for explainability annotations, and one to choose the best sequence of word labels (i.e., which words are important and which are not). 

For the former component, exhaustively generating all possible label assignments is prohibitively expensive (i.e., $O(2^N)$ for a sequence of length $N$). To mitigate this cost, we rely on the prediction scores of the EC classifier to reduce the number of candidates. That is, if the score of the binary EC for a given token is higher than a threshold ($t_{up}$), we directly annotate the corresponding token as important (i.e., assign label 1); if this score is lower than a second threshold ($t_{low}$), we annotate the token as not important (label 0); and, lastly, if the the score is between the two thresholds, we generate two candidate labels for this token (both 0 and 1). For example, given an input sentence {\em ``[CLS] [SUBJ-PER] was born in [OBJ-CITY] .''},\footnote{The entities participating in a relation are masked with their named entity labels (see Section~\ref{sec:models}).} and these prediction scores from the EC: {\tt [0.12, 0.14, 0.19, 0.86, 0.25, 0.15, 0.01]}, using  $t_{up} = 0.8$ and $t_{low} = 0.2$, we produce the following candidate label sequences: {\tt [0, 0, 0, 1, 0, 0, 0]} and {\tt [0, 0, 0, 1, 1, 0, 0]}, because the assignment for the token {\em in} is ambiguous according to the two thresholds. 

Once these candidates are generated, we loop through all the generated sequences of word labels, and pick the sequence $\hat{c}$ that yields the highest score for the correct relation label according to the current RC:

\begin{small}
\begin{equation}
    \hat{c} = \argmax_c p(R|c)
\end{equation}
\end{small}
where $R$ is the gold label of the instance, $p(R|c)$ is the score at the gold label $R$ predicted by the RC for a given annotation candidate $c$.
In the previous example, if the RC scores of the two candidates for the correct relation label  {\tt per:city\_of\_birth} are 0.8 and 0.5, we select the first candidate over the second one.

Then this sequence of labels is used as (pseudo) gold data to train the EC on this training example. This guarantees that each training example has annotations (gold, or generated through the above procedure) for both EC and RC. 

Because these two components rely on having reasonable predictions from the EC and RC classifiers, we found it beneficial to include the previous burn-in period, where these classifiers are trained using the (small) amount of supervision available. 

\subsubsection{Explainability Annotation}
\label{sec:expann}
As mentioned, a key part of our approach requires that EC annotations be available for a few of the training examples. To this end, rather than relying on manual annotations, which are expensive, we repurpose rules that extract the same relation. 
The intuition behind our approach is that if a rule exists that extracts the same relation label as the gold label in a training example, then this rule (and, specifically, its lexical elements) can be seen as an explanation of the extraction. 
In particular, in this article we focus on the TACRED dataset~\cite{zhang2017tacred}, and select explanations from two sets of rules:

%

{\flushleft {\bf (1) Surface rules:}} The TACRED project generated a set of high-precision rules for the task, implemented in the Tokensregex language~\cite{chang2014tokensregex}. For example, the rule {\tt SUBJ-PER was born in * OBJ-CITY}\footnote{We simplified the Tokensregex syntax for readability.} extracts a {\tt per:city\_of\_birth} relation between a person named entity (the subject) and a city named entity (the object) if the sequence {\em was born in} occurs somewhere between the two entities. For such rules, we label all tokens contained in the rule (e.g., {\em was}, {\em born}, {\em in}) with the label 1 (i.e., they are important for explainability), and all other tokens in the sentence with 0. 

\begin{figure}[t]
\begin{lstlisting}[style=odin-style, escapechar=;]
;\colorbox{yellow!12}{\textbf{label}: per:employee\_of};
pattern: |
 trigger = 
    ;\colorbox{blue!12}{[lemma=/work|write|play|consult|serve/]};
 ;\colorbox{red!12}{$\,$\textbf{\textcolor{blue}{subject}}:$\,$SUBJ\_Person}; = <acl? nsubj
 ;\colorbox{red!12}{$\,$\textbf{\textcolor{blue}{object}}:$\,$OBJ\_Organization}; = nmod
\end{lstlisting}

\begin{tabular}{l p{7cm}}
\cellcolor{yellow!12} \hspace{0.1cm} & {Label(s) to assign to a match.} \\[.3ex]
\cellcolor{blue!12} \hspace{0.1cm} & {Lexical constraints on the relation's predicate.} \\[.3ex]
\cellcolor{red!12} \hspace{0.1cm} & {\texttt{argName:ArgType}, where \texttt{ArgType} indicates the named-entity category expected for this argument.} \\
\end{tabular}
\caption{An example of a relation extraction rule in the Odin language that extracts the {\tt per:employee\_of relation} relation. The rule is driven by verbal triggers such as  {\em work, play} or {\em serve}. The relation's arguments (the subject and object) are identified through both semantic constraints (subject must be {\tt Person}), and syntactic ones (subject must be attached to the trigger through a certain syntactic dependency pattern: an optional ({\tt ?}) adnominal clause ({\tt acl}), followed by a nominal subject ({\tt nsubj}). 
This rule would extract a \texttt{per:employee\_of} relation from the text {\em ``\dots Joe is a research scientist working at IBM\dots''}.
}
\label{fig:rule}	
\end{figure}

{\flushleft {\bf (2) Syntactic rules:}} In initial experiments, we observed that the TACRED surface rules have high precision but low recall. To improve generalization, we also wrote 38 syntax-based rules using the Odin language~\cite{valenzuela-escarcega-etal-2016-odins}.\footnote{All these rules are included in this submission as supplemental material.}
Figure ~\ref{fig:rule} shows an example of such a rule. 
For these syntactic rules, we marked all their lexical elements (typically the trigger predicates such as {\em work} or {\em write} in the figure) as important (label 1), and all other words as not important (label 0).

\subsubsection{Classifiers}
\label{sec:models}
\begin{figure*}[t]
    \centering
    \includegraphics[width=\textwidth]{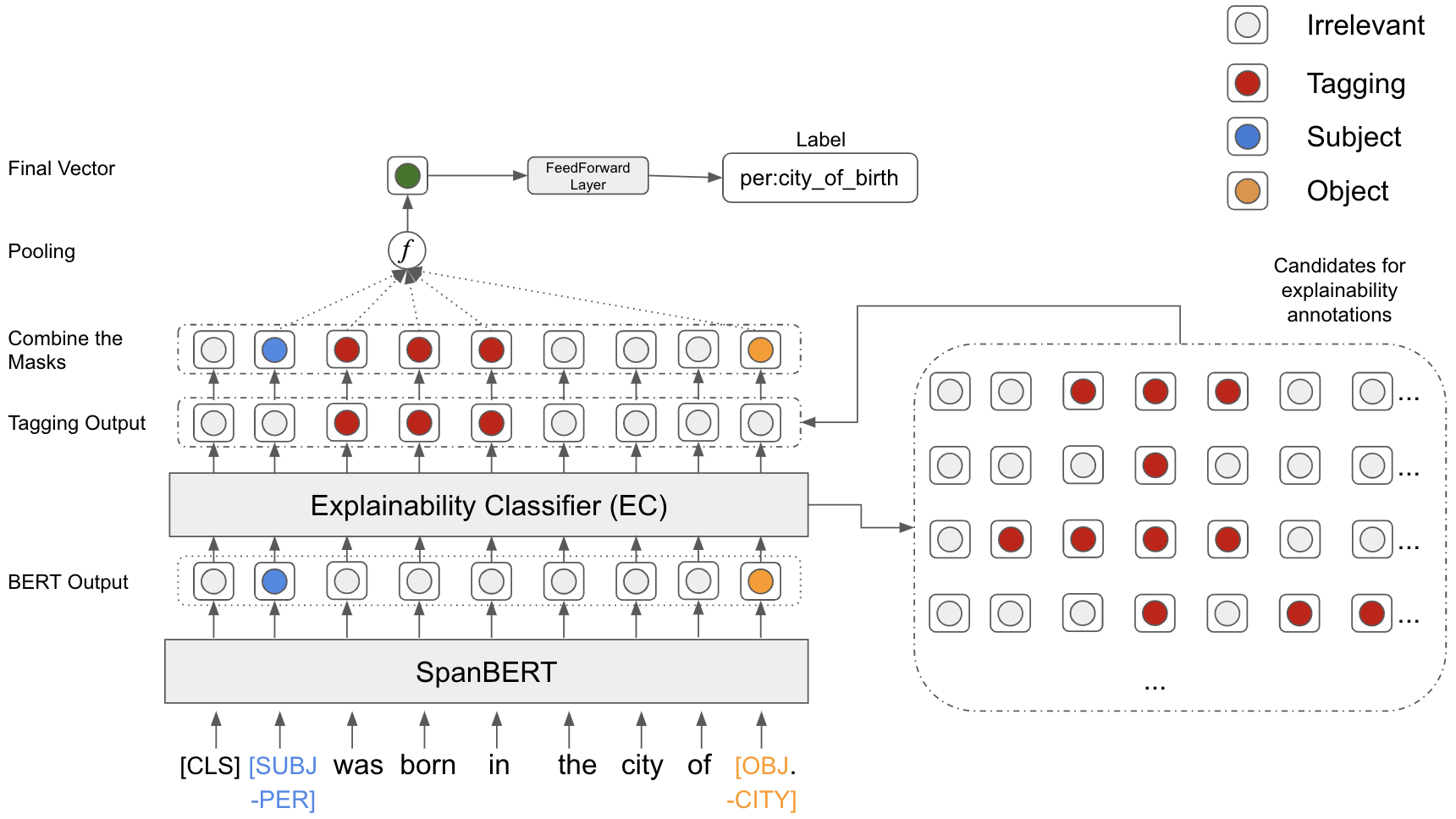}
    \caption{Neural architecture of the proposed multitask learning approach. The entity tokens (subject in \textcolor{blue}{blue} and object in \textcolor{orange}{orange}) are masked with their named entity labels, e.g., {\tt SUBJ-Person}, in the actual implementation. 
    }
    \label{fig:architecture}
\end{figure*}

As mentioned, the building blocks of our approach consist of three classifiers: the no-relation classifier (NRC), the relation classifier (RC), and the explainability classifier (EC). These are jointly trained using the schema previously described in this section. Below we describe their individual details, which are also visualized in Figure~\ref{fig:architecture}.


\paragraph{SpanBERT Encoder and NRC:}
We follow the entity masking schema from ~\cite{zhang2017tacred} and replace the subject and object entities with their provided named entity (NE) labels, e.g., ``{\em [CLS] [SUBJ-PER] was born in [OBJ-CITY]} \dots''. We feed this input to a SpanBERT-based ~\cite{joshi-etal-2020-spanbert} encoder:
\begin{equation}
   [\pmb{h}_{0},\ldots,\pmb{h}_{n}] = \textrm{Encoder}([w_{0},\ldots,w_{n}]) 
\end{equation}
where $w_n$ is the id of the word at position $n$, and $h_n$ is the hidden representation generated by the encoder.
We add the special masking tokens for {\tt SUBJ--*} and {\tt OBJ--*} to the vocabulary so that the encoder can handle them properly. 
We implement the NRC using a feedforward layer with a sigmoid function on top of the encoder's {\tt [CLS]} token.

\paragraph{Explainability Classifier (EC):}
We implement the EC as a binary token-level classifier, where the positive label indicates that the corresponding token is important for the underlying relation.
Section~\ref{sec:expann} discusses how these annotations are generated from rules; Section~\ref{sec:training} explains the SSL training procedure when these annotations are not available.

\paragraph{Relation Classifier (RC):}
Crucially, the RC relies {\em only} on words that are marked as important by the EC, or are part of the subject/object entity.
This is an important distinction between our approach and other relation extraction methods, which typically rely on the {\tt [CLS]} representation for classification.
In the next section, we empirically show that this latter strategy is considerably less explainable than ours. This is because the {\tt [CLS]} representation aggregates information from all tokens in the sentence, whereas our method focuses only on the important ones. 

We build the aggregated representation of the important context words, subject and object as follows:
\begin{equation}
   \pmb{h}_{final}= f(\pmb{h}_{ctx_1:ctx_n}) \circ f(\pmb{h}_{subj_1:subj_n}) \circ f(\pmb{h}_{obj_1:obj_n})
\end{equation}

where $\pmb{h}$ denotes the hidden representations produced by the encoder, $f:\mathbb{R}^{d \times n} \rightarrow \mathbb{R}^{d}$ is the average pooling function that maps from $n$ output vectors into one; and $\circ$ is the concatenation operator. Importantly,  $\pmb{h}_{ctx}$ iterates only over words marked as important by the EC. 

The concatenated representation $\pmb{h}_{final}$ is fed to a feedforward layer with a softmax function to produce a probability distribution $\pmb{p}$ over relation types. 

The three classifiers are trained using the following joint loss function:
\begin{equation}
    loss = loss_{nrc} + loss_{ec} + loss_{rc}
\end{equation}
\begin{equation}
    loss_{nrc} = -(t^{n}*log(y^n) + (1-t^{n})*log(1-y^n))
\end{equation}
\begin{equation}
    loss_{ec} = -(t^e*log(y^e) + (1-t^e)*log(1-y^e))
\end{equation}
\begin{equation}
    loss_{rc} = -log(p(R))
\end{equation}
where the losses of the NRC and EC ($loss_{nrc}$ and $loss_{ec}$, respectively) are implemented using binary cross entropy. For both, $t$ indicates the corresponding gold label, and $y$ is the respective sigmoid's activation. The loss of the RC ($loss_{rc}$) is implemented using categorical cross entropy, where $p(R)$ is the likelihood predicted by the model for the correct relation $R$.

%

\subsection{Aggregating Local Explanations into a Global, Rule-based Model}
\label{sec:global}

As mentioned, the last component of our approach aggregates all local RC and EC predictions into a single rule-based model that explains the overall behavior of the RC and EC models. As such, the produced rule-based model brings global explainability to the task. We will show in Section~\ref{sec:experiments} that this transformation comes with a cost in performance, but this cost might be acceptable in scenarios where such RE extraction must be deployed, maintained, and improved over a long period of time.



\subsubsection{Rule Generation}
As shown in Table \ref{tab:walk} (c), our relation and explanation classifiers produce all the information necessary to generate an Odin rule. At a high-level, the Odin rules we employ here follow a predicate (or {\tt trigger} in the Odin language) and argument template, where all arguments are connected to the trigger using a syntactic dependency path. This information is either provided by our classifiers, e.g., we use the rationale tokens identified by the EC as triggers, or can be automatically extracted from the sentence, e.g., we represent the syntactic connections between predicate and arguments using the shortest path that connects them in the syntactic dependency tree. Algorithm~\ref{alg:rule_gen} describes this entire rule generation process:

\begin{algorithm}[H]
	\caption{Rule Generator}
	\begin{algorithmic}[1]
	\Require set of annotated sentences, $\mathbb{S}$
	\Require model output $\mathbb{L}$ (from RC) and $\mathbb{T}$ (from EC)
	\State $\mathbb{R} \gets \emptyset$
		\For {every sentence $s$ in $\mathbb{S}$}
		\State Get the subject and object entity $e_s$ and $e_o$ from $s$
		\State Get the predicted relation label $l$ from $\mathbb{L}$ and the rationale words $t$ from $\mathbb{T}$
		\If{ $s$ hasn't been extracted by any manual rule}
		        \State Find the shortest path $p_s$ between $t$ and $e_s$ in the dependency tree
		        \State Find the shortest path $p_o$ between $t$ and $e_o$ in  the dependency tree
				\State $r \gets$ empty Odin rule template
				\State Assign $l$ to $r$ as the label to match
				\State Assign $t$ to $r$ as the relation predicate (or trigger)
				\State Assign $p_s$ and $p_o$ to $r$ as the argument patterns.
				\State $\mathbb{R} \gets \mathbb{R} \cup \left\{ r \right\} $	
		\EndIf		
		\EndFor
		\Ensure set of generated rules $\mathbb{R}$
	\end{algorithmic}
	\label{alg:rule_gen}
\end{algorithm}

%% file: sections/experiments.tex
\section{Experimental Results}
\label{sec:experiments}



\subsection{Data Preparation}
\label{sec:dataprep}
We report results on the TACRED dataset~\cite{zhang2017tacred} and CoNLL04 dataset~\cite{roth-yih-2004-linear}. As discussed in Section~\ref{sec:expann}, we provided rules for explanation supervision. For the TACRED data, we selected rules from the surface patterns of~\citet{Angeli2015BootstrappedST}, and we combined them with an additional set of 38 syntactic rules in the Odin language~\cite{valenzuela-escarcega-etal-2016-odins} that were manually created by one of the authors from the training data. For CoNLL04 data, we selected from a set of 19 syntactic rules in Odin language, 10 of which are borrowed from the TACRED syntactic rules, since the two datasets shared some overlapping relations.

These rules match 20.7\% of positive examples in the TACRED training set and 24.2\% of positive examples in the CoNLL04 training set. On average, 7.27 rules are assigned to each TACRED relation, and 3.8 rules are assigned to each CoNLL04 relation.

Importantly, our approach does {\em not} use rules at evaluation time. However, we take advantage of all existing rules to automatically evaluate the quality of the explanations generated by our method. In the TACRED dataset, the combined set of rules from~\cite{Angeli2015BootstrappedST} and our syntactic rules match 23.9\% data points in the development set, and 23.9\% examples in the test set; in the CoNLL04 dataset, the syntactic rules match 20.1\% of examples and 20.9\% of examples respectively. We use only these matches for an automated evaluation of explainability (discussed below).

\subsection{Baselines}
\label{sec:baselines}
\subsubsection{Relation Extraction Baselines}
\label{sec:baselines_rc}
For the relation extraction task, we compare our approach with three baselines: an extended version of the rule-based approach of~\mbox{\citet{Angeli2015BootstrappedST}}, a neural state-of-the-art RE approach based on SpanBERT~\mbox{\cite{joshi-etal-2020-spanbert}}, and a neural approach with built-in explainability~\cite{lei-etal-2016-rationalizing}:

\begin{itemize}
\item \textbf{Rule-based Extraction.} As mentioned in Section \ref{sec:dataprep}, we employ two sets of rules. First, we use the tokensregex surface rules from \cite{Angeli2015BootstrappedST}, which are executed in the Stanford CoreNLP pipeline~\cite{manning-etal-2014-stanford}. Second, we include the Odin syntactic rules we developed in-house, which are executed in the Odin framework~\cite{valenzuela-escarcega-etal-2016-odins}.\footnote{The rule set from \cite{Angeli2015BootstrappedST} also included some syntactic rules, but we found out that they only matched the simpler {\tt per:title} relation, so we did not use them.}

\item \textbf{SpanBERT.} SpanBERT~\cite{joshi-etal-2020-spanbert} is an extension of the original BERT~\cite{devlin2018bert} that: (1) masks continuous random spans instead of random tokens, and (2) trains the span boundary representations to predict the full content of the masked span without depending on individual token representations within it. SpanBERT outperforms BERT in many tasks including relation extraction. Further, SpanBERT is currently the best TACRED BERT-based model available in the HuggingFace transformer library~\cite{wolf-etal-2020-transformers} that does not use any external resources, or does not rely on complex hybrid architectures.

\item \textbf{Unsupervised Rationale.} \citet{lei-etal-2016-rationalizing} proposed an approach that combines an {\em unsupervised} rationale generator with a task-specific classifier, both of which are trained to operate together (similar to our approach). However, there are several key differences between their method and ours. First, their explanation generator cannot incorporate human input (as we do through rules); instead, it is indirectly guided by the loss of the downstream task. Second, their architecture is more complex, i.e., they use two distinct encoders: one for explanation generation and another for the downstream task (both of which are implemented with recurrent networks). 
We adapt this method to our RE framework, by replacing our EC with their rationale generation algorithm (which is a token-level binary classifier that produces an output compatible with our EC). For a fair comparison with our method, we kept the other components unchanged. That is: 
we encode the input text using the same SpanBERT, then we use their generated rationales and the given entities as pooling mask to construct the final vector to feed into the relation classifier\footnote{We also observed that our architecture that uses a single, shared transformer encoder performs better than their original architecture with two distinct encoders.}. Originally, \citet{lei-etal-2016-rationalizing} proposed their approach to sentiment analysis and text retrieval. \citet{bastings-etal-2019-interpretable} extended this method and adapted it to a natural language inference task.
To our knowledge, this is the first attempt to apply this explainability strategy to relation extraction.

\end{itemize}
Note that all baselines as well as our method receive inputs in the standard TACRED format,\footnote{We converted the CoNLL04 data into the same format as TACRED.} which contains tokenized sentences, spans of the subject and object mentions, and the types of the two entity mentions.
The only difference between the RC baselines and our method is that, as discussed in Section~\ref{sec:dataprep}, our approach receives information on which sentence tokens were matched by rules during the burn-in training period.


\subsubsection{Explainability Baselines}
For explainability, we compare our approach against eight baselines, detailed below. 
These are all popular explanation approaches published in recent years. 
Most of them provide a feature importance score for each feature\footnote{Except for greedy adding and unsupervised rationale approaches which rely on labeling the features to be included in the rationale, similar to what we do.} and most of them are post-hoc\footnote{Except for unsupervised rationale approach which trains a generator together with the rest of the model, similar to what we do.}. 
Here, we labeled the top $N$ positive features identified by the baselines as important.\footnote{We ignored tokens part of the subject and object entities for a fair comparison.} In the first quantitative evaluation of explainability (Section~\ref{sec:quanteval}), for all baselines we set $N$ to be equal to the number of words in the gold explanation. Importantly, this means that all baselines have an unfair advantage over our approach, which is non-parametric with respect to $N$, i.e., it identifies $N$ on the fly for each sentence. In the second, qualitative evaluation of explainability (Section~\ref{sec:qualeval}), $N$ is a hyper parameter that we tuned to maximize the baselines' performance.\footnote{We used $N=3$ for TACRED, and $N=1$ for CoNLL04.} 

We detail the eight explainability baselines below:

\begin{itemize}
  \item \textbf{Attention.} Attention weights have been proposed as an explanation mechanism by~\citet{bahdanau2014neural}. Followup work debated the validity of this strategy~\cite{ jain2019attention, wiegreffe2019attention, kobayashi-etal-2020-attention}. However, because this remains a popular approach, we include attention weights as a baseline in this work. In particular, we use the attention weights from the last layer of a ``vanilla''  SpanBERT model, i.e., one that is trained on top of the {\tt [CLS]} representation, without an EC. For this baseline, we label as important the top $N$ tokens with the highest {\tt [CLS]} attention weights. 
  \item \textbf{Saliency Mapping.} The feature importance score of the token $x_i$ is determined by the highest prediction's accumulated gradients in each dimension of the token in the embedding layer. These scores are obtained through a back-propagation of the highest prediction's probability. Although there are different implementations of the gradient saliency mapping approach~\cite{devlin2018bert, voita2021analyzing}, we use the simple back-propagation approach from \cite{simonyan2013deep}.
  \item \textbf{LIME.} \citet{ribeiro2016should} proposed the LIME framework, which provides explanations to any black-box classifier. LIME samples the neighbors of the local instance $\bm{x}$ to be explained, by generating perturbations of the tokens in $\bm{x}$. Then, it trains a linear separator from these samples to approximate the local behavior of the model. The coefficients of the separator are later used as the feature importance score.
  \item \textbf{Unsupervised Rationale.} As mentioned in the previous sub-section, this baseline replaces our EC with the unsupervised method of \citet{lei-etal-2016-rationalizing}. Here we use this method as an explainability baseline.
  \item \textbf{SHAP.} The Shapley value~\cite{P-295} is a cooperative game theory concept that calculates the score of feature $x_i$ by taking into account its interactions with all other subsets of features. Similar to what LIME does, \citet{NIPS2017_7062} also train a linear model to approximate the local behavior around the sampled neighbors. However, unlike LIME, which uses cosine similarity or L2 distance as its kernel, they propose a SHAP kernel which is determined by the number of permutations of features. 
  \item \textbf{CXPlain.} \citet{schwab2019cxplain} proposed an approach called CXPlain that explains the decisions of any machine-learning model by measuring the importance of the model's features. To this end, CXPlain masks each token $x_i$ in $\bm{x}$, and calculates the score of $x_i$ by comparing the output with the masked input $\overline{\bm{x}}$ against the output that relies on the original input $\bm{x}$. The difference between the two is calculated using a causal objective.
  \item \textbf{Greedy Adding.} Instead of randomly sampling from perturbations or masking the features, \citet{vafa2021rationales} proposed a method that greedily adds the features to the input data point. That is, it starts with an empty rationale, and each time it selects and adds the feature that increases the probability of the correct label $y_t$ the most. The process repeats as long as the confidence in predicting $y_t$ keeps increasing.
	\item \textbf{All Words between Subject and Object.} We have observed that most of the important words that determine the relation between the entities occur in the span between the two entities. To capture this intuition, we implemented this simple baseline, which simply includes all the words between subject and object in its rationale.
\end{itemize}
Similarly to the RC settings discussed in the previous sub-section, these baselines and our method rely on the standard TACRED input format. However, our EC is semi-supervised, i.e., during burn-in it receives explainability annotations generated by rules. In contrast, the EC baselines do not rely on rule information. 

\subsection{Implementation and Evaluation Details}
Before introducing our results, we discuss key details about our implementation and evaluation.

To avoid the RC classifier overfitting on the names in the sentence~\cite{suntwal-etal-2019-importance}, 
we mask the subject and object entities by replacing the original tokens in these entities with a special token, i.e., {\tt SUBJ--<NE>} or {\tt OBJ--<NE>},  where {\tt <NE>} is the corresponding name entity type provided in the dataset.
We use the pre-trained SpanBERT to encode the input sentence. For the TACRED dataset, which is organized to contain a single relation per sentence,
 we feed the {\tt [CLS]} token to the final linear layer for relation classification. However, for the CoNLL04 data, which typically contains more than one relation per sentence, we used the concatenation of  the {\tt [CLS]} hidden state and the average pooling of  {\tt [SUBJ]} and {\tt [OBJ]} hidden state embeddings. This was necessary to distinguish between the different relations that co-occur in the same sentence. 
We used the AdamW optimizer~\cite{loshchilov2019decoupled} for all training processes. 
We evaluated all RC classifiers using the standard micro precision, recall, and F1 scores. 
All neural models were trained using 5 different random seeds; we report the average scores and standard deviation over these seeds for RC.

For explainability, we report two evaluations.\footnote{We did not include an evaluation of faithfulness, which is typically done by post-hoc explainability approaches~\cite{ribeiro2016should, schwab2019cxplain} because our approach is faithful by design, i.e., our RC only relies on the tokens identified by the EC.} For the first, automated evaluation, we use only the data points that are associated with a rule that produces the same relation label as the gold data. For these examples, we consider the lexical artifacts of the rule as gold information for explainability (as explained in \S\ref{sec:expann}). We measure the overlap between the important words produced by the analyzed methods and this data using precision, recall, and F1 scores. 
We also include a second, qualitative evaluation on the {\em plausability} of the generated explanations \cite{vafa2021rationales}, where a more plausible explanation will overlap more with a relation explanation manually generated by domain experts. 
For this evaluation, we sampled 100 and 60 data points from the test sets of TACRED and CoNLL04, respectively. These are sentences where our model predicted a relation, and where there is {\em no} gold annotation from rule-based method (i.e., no rule matched). We split these data points into two sets: a subset where our method predicted the correct relation, and one where it did not. In other words, in the former set, we investigate the capacity of the explainability methods to explain correct predictions, while in the latter we analyze their capacity to explain why the machine was incorrect.
Two domain experts\footnote{These were two of the authors.} manually annotated rationales for these sentences and the provided relation labels. The annotators were asked to identify the minimal set of tokens that explain the provided relation. Or, in other words, identify the tokens that when replaced with other words change the relation to be predicted. For example, in the sentence {\em SUBJ-PER was born in OBJ-CITY.}, if we replace the words {\em born in} with other words (e.g., {\em moved to}), the relation between the subject and object changes.
Importantly, to avoid any potential bias, the two annotators worked completely independently of each other, and had no access to explanations provided by any algorithm.\footnote{To encourage reproducibility, we release the annotations at \url{https://github.com/clulab/releases/tree/master/cl2022-twoflints/dataset} }
We evaluate the overlap between the machine and human rationales using the same standard precision, recall, and F1 measures. 


Appendix A lists the hyperparameters used to train all RC and EC models. 

Lastly, we evaluate the quality of the generated rule-based model. 
To this end, we evaluated two sets of rules: rules generated from the training sentences,\footnote{We filter our training relations which matched a gold rule, since there is already a rule assigned to them} and rules generated over the test set. In the latter scenario, we do not use any gold data. That is, we rely on the predicted relation labels (from the RC) and rationales (from the EC) to generate rules. 
Thus, the latter setting is akin to transductive learning, i.e., where the model has access to the unlabeled data from the testing partition, but no access to any human annotations. 
We evaluate the performance of these rule-based models using the same micro precision, recall, and F1 scores as the first RC evaluation.


\subsection{Results and Discussion}
In this section, we introduce and discuss the results for both relation and explainability classification. We conclude this section with an error analysis that highlights some typical errors in our models.

\subsubsection{Relation Extraction}
\label{sec:rc}

\begin{table*}[]
\centering
\begin{tabular}{rccc}
Approach            & Precision & Recall & F1    \\ \hline
\multicolumn{4}{c}{Baselines} \\
\hline
Rules  &  {\bf 85.82} & 24.21         & 37.77         \\
SpanBERT~\cite{joshi-etal-2020-spanbert} & 69.97{\small$\pm 0.58$}      & {\bf 70.20{\small$\pm 1.73$}}    &  70.07{\small$\pm 0.73$} \\
Unsupervised Rationale & 69.24{\small$\pm 0.40$} & 69.05{\small$\pm 1.86$}  &   69.14{\small$\pm 0.83$}    \\
\hline
\multicolumn{4}{c}{Our Approach} \\
\hline
Burn-in Only &   51.06{\small$\pm 3.57$} & 48.32{\small$\pm 2.33$}  &   49.61{\small$\pm 2.42$}    \\
Full Model & 72.02{\small$\pm 0.90$}     & 69.11{\small$\pm 1.82$} &  {\bf 70.52{\small$\pm 0.54$}} \\
\end{tabular}
\caption{ Relation extraction results on the TACRED test partition. We used the pre-trained SpanBERT-large. 
Our full model trains on the entire training partition using the SSL method discussed in Section~\ref{sec:training}. 
The ``burn-in only'' setting trains just on the training subset that has annotations from rules.
}
\label{results_1_tac}
\end{table*}

\begin{table*}[]
\centering
\begin{tabular}{rccc}
Approach            & Precision & Recall & F1    \\ \hline
\multicolumn{4}{c}{Baselines} \\
\hline
Rules   &81.6 & 16.82         & 27.90        \\
SpanBERT~\cite{joshi-etal-2020-spanbert} & 81.30{\small$\pm 4.89$}      &71.01{\small$\pm 5.11$}   &75.78{\small$\pm 4.79$} \\
Unsupervised Rationale & {\bf 83.91{\small$\pm 2.88$}} & 74.88{\small$\pm 1.44$}  &   79.11{\small$\pm 1.01$}   \\
\hline
\multicolumn{4}{c}{Our Approach} \\
\hline
Burn-in Only&  62.71{\small$\pm 2.27$} & 53.32{\small$\pm 0.95$}  &   57.63 {\small$\pm 1.39$}   \\
Full Model &   83.01{\small$\pm 2.16$}    &{\bf 76.30{\small$\pm 3.08$}} & {\bf 79.46{\small$\pm 0.92$}} \\
\end{tabular}
\caption{ Relation extraction results on the CoNLL04 test partition. We used the pre-trained SpanBERT-large. 
Our full model trains on the entire training partition using the SSL method discussed in Section~\ref{sec:training}. 
The ``burn-in only'' setting trains just on the training subset that has annotations from rules.
}
\label{results_1_conll}
\end{table*}

Tables~\ref{results_1_tac} and~\ref{results_1_conll} report the RE performance of all methods discussed on the TACRED and CoNLL04 datasets. The results of all statistical approaches are averaged over three random seeds. For all these models we report average performance and standard deviation in the tables. We draw the following observations from these tables:
\begin{itemize}
\item First, the SSL variant of our approach improves considerably over the equivalent burn-in only setting (i.e., training just on the data points that have matching rules). The improvement is 20.91\% F1 (absolute) on TACRED, and 21.83\% (absolute) on CoNLL04. These results highlight the importance of SSL for this task.
\item Second, our approach is slightly better than SpanBERT on TACRED, and yields a statistically-significant improvement of nearly 4\% F1 (absolute) on CoNLL04.\footnote{We performed statistical significance analysis using non-parametric bootstrap resampling with 1000 iterations.} This indicates that  jointly training for classification and explainability helps the classification task itself (or, in the worst case, does not hurt relation classification). Table~\ref{results_1_conll} also shows that our approach has the highest RE recall on CoNLL04, higher than the vanilla SpanBERT by 5\%. 
All in all, this suggests that explainability also serves as a disambiguator in situations where multiple relations co-occur in the same sentence (the common setting in CoNLL04) by narrowing the text to just the context necessary for the relation at hand. As further evidence that performing RC on top of explanations helps disambiguate the underlying text, the standard deviation of our approach on CoNLL04 is five times smaller than that of SpanBERT. 

\item Interestingly, the unsupervised rationale method approaches the performance of our full model on both datasets. However, as we will show in the next sub-section, this comes with considerably worse explanations.

\item Lastly, our approach nearly doubles the F1 score of the rule-based approach on TACRED, and more than doubles it on CoNLL04. This is caused by large improvements in recall, which highlights the importance of hybrid strategies that combine rules and neural components. 
\end{itemize} 

To understand the runtime overhead introduced by the EC, we compared our method's runtimes during training and inference against the runtime of the vanilla SpanBERT.
The average training time of our method is 0.37 sec/batch in the burn-in period and 0.38 after burn-in. In contrast, the average training time of SpanBERT is 0.06 sec/batch.\footnote{All times measured on an NVIDIA RTX 3090 GPU.} The inference time for both our model and SpanBERT is 0.10 sec/batch on the same device. The larger overhead in training is caused by: (a) back-propagating through a larger computational graph due to the joint EC and RC loss, and (b) iterating through multiple candidate explanations. We measured the average number of explanation candidates to be 85 in the first training epoch after burn-in period, and 22 after 10 epochs.
However, considering that inference time are similar, we believe that the training overhead is justified by the additional explainability functionality included in the framework.




\subsubsection{Quantitative Evaluation of Explainability}
\label{sec:quanteval}

\begin{table*}[t!]
\centering
\begin{tabular}{rccc}
Approach            & Precision & Recall & F1    \\ \hline
Attention &30.28&30.28&30.28\\
Saliency Mapping    &  30.22&  30.22&30.22 \\
LIME &      30.45&36.84&  32.49\\
Unsupervised Rationale & 4.65 & 79.53 & 8.51 \\
SHAP&    31.27 &  31.27 &  31.27\\
CXPlain &     53.60&  53.60& 53.60\\
Greedy Adding& 40.47     &   50.53   &  40.81\\
All words in between SUBJ \& OBJ    &  71.48 & 86.33   &   78.21   \\
Our Approach & \bf{95.63}     & \bf{97.92}   & \bf{95.76}  \\
\end{tabular}
\caption{ Automated evaluation of explainability on TACRED, in which we compare explainability annotations produced by these methods against the lexical artifacts of rules. 
}
\label{results_2_tac}
\end{table*}

\begin{table*}[t!]
\centering
\begin{tabular}{rccc}
Approach            & Precision & Recall & F1    \\ \hline
Attention &69.44&69.44&69.44\\
Saliency Mapping    &  42.42&  42.42&42.42 \\
LIME &              62.45&89.39&  68.45\\
Unsupervised Rationale & 5.47 & 86.94 & 9.84 \\
SHAP&     34.85 & 34.85 &  34.85\\
CXPlain &  50.00 & 50.00 & 50.00  \\
Greedy Adding&  23.24&   54.55& 29.58\\
All words in between SUBJ \& OBJ    & 72.99 & 96.59  &  77.29 \\
Our Approach & \textbf{99.29}     & \textbf{100}   & \textbf{99.52}  \\
\end{tabular}
\caption{ Automated evaluation of explainability on CoNLL04, in which we compare explainability annotations produced by these methods against the lexical artifacts of rules. 
}
\label{results_2_conll}
\end{table*}

The results of the automated evaluation of explainability in Tables~\ref{results_2_tac} and~\ref{results_2_conll} show that our approach generally improves explainability quality considerably. Post-hoc explanation methods do not provide the same explanation quality compared to our method, which actively models explainability. Note that the high performance of annotating all the words between subject and object is caused by the fact that most data points in this evaluation are associated with surface rules, which prefer shorter contexts that are more likely to contain only significant information.
Nevertheless, the 20\% F1 gap between this strong baseline and our method indicates that our method successfully learns how to generalize beyond these simple scenarios. 

However, we note that these results are not terribly surprising: our method is trained to generate explanations that mimic lexical artifacts of rules, while the other explainability baselines have not been exposed to rules during their training. Thus, this evaluation is necessary (to validate that our approach is learning to do what we intended, which is to mimic the lexical artifacts of rules) but not sufficient. In the next sub-section, we will show that our approach overlaps with human explanations much more than all other explainability baselines. 

\begin{table}[t!]
\centering
\begin{tabular}{rccc}
Num of Rules             & Precision & Recall & F1 \\ \hline
\multicolumn{4}{c}{Relation Classification} \\ \hline
Up to top 1 (0.98 rules/relation)    &  72.48 & 66.23 & 69.21\\
 Up to top 5 (3.56 rules/relation)    &  {\bf 72.97} & 69.02 & 70.94\\
Up to top 10 (5.02 rules/relation)  &   69.30 & 71.64 & 70.45\\
All rules (7.27 rules/relation)        &  71.15 & {\bf 71.13}&  {\bf 71.14}\\
\hline \multicolumn{4}{c}{Explainability Classification} \\ \hline
Up to top 1 (0.98 rules/relation)    & 74.62 & 85.35 & 75.02\\
Up to top 5 (3.56 rules/relation)    & 92.19 & 94.06 & 91.28\\
Up to top 10 (5.02 rules/relation)  & 91.06 & 95.62 & 91.22\\
All rules (7.27 rules/relation)        &  \bf{95.63}     & \bf{97.92}   & \bf{95.76}\\
\end{tabular}
\caption{ Learning curve of our approach on TACRED based on amount of rules used. In each experiment, we use up to top $k$ rules per relation type; the number in parentheses is the actual average number of rules per type. 
}
\label{results_4}
\end{table}

Table~\ref{results_4} lists a learning curve for our approach on TACRED, as we vary the amount of rules available per relation. That is, for each relation, we use up to top $k$ rules, where $k$ varies from 1 to 10. In the table we include results for both relation and explainability classification using the same measures as the previous tables. The table shows that even in the ``up to top 5 rules'' configuration (which means an average of 3.6 rules per relation type in practice), our model obtains a close F1 to the our best model with good explainability. This result indicates that our approach performs well with minimal human supervision for explanation guidance. Note that we do not include the learning curve for CoNLL04 since there are only 19 rules applied to this dataset, which translates into only 3.8 per relation type. 

\subsubsection{Qualitative Evaluation of Explainability}
\label{sec:qualeval}

 \begin{table}[]
\centering
\begin{tabular}{rccc}
Approach            & Precision & Recall & F1    \\ \hline
Attention &41.39  & 20.60 & 26.50\\
Saliency Mapping    &  18.73 &35.58 &23.41 \\
LIME &     14.31 & 26.03 &  18.09\\
Unsupervised Rationale & 4.73 & 69.66 & 8.30 \\
SHAP&    13.86 & 22.85  &  16.79\\
CXPlain &    28.84 & 55.06 & 36.48\\
Greedy Adding&  31.59    &  33.52   &  30.16\\
Our Approach & \textbf{74.72}     & \textbf{61.20}   & \textbf{62.05}  \\
\end{tabular}
\caption{TACRED evaluation of the plausability of explanations, which measures the overlap between machine explanations and human annotations. For each method, we pick the higher scores between the two human annotators.
}
\label{results_human_tacred}
\end{table}

 \begin{table}[]
\centering
\begin{tabular}{rccc}
Approach            & Precision & Recall & F1    \\ \hline
Attention & 61.06 & 30.30 & 38.94\\
Saliency Mapping    & 18.79 &  39.39&24.43\\
LIME &     22.14 &53.33&  30.09\\
Unsupervised Rationale & 5.35 & 74.55 & 9.31 \\
SHAP&   18.18  &36.36   &  23.27\\
CXPlain &   21.21  & 44.55& 27.82\\
Greedy Adding&  33.33 & 38.03     &  32.21\\
Our Approach & \textbf{65.15}     & \textbf{59.24}   & \textbf{58.97}  \\
\end{tabular}
\caption{CoNLL04 evaluation of the plausability of explanations, which measures the overlap between machine explanations and human annotations. For each method, we pick the higher scores between the two human annotators.
}
\label{results_human_conll}
\end{table}
 
Tables~\ref{results_human_tacred} and~\ref{results_human_conll} lists the results of our evaluation of the plausability of explanations by comparing them against human annotations of explainability. Similar to evaluations of machine translation, we choose the higher scores between the machine methods and any of the two human annotators. Note that the human annotators had a Kappa agreement \cite{mchugh2012interrater} of 69.8\% on labeling the same tokens as part of an explanation. This is considered moderate~\mbox{\cite{landis1977measurement}}, which we found encouraging considering the complexity of the task and the fine granularity of the annotations. We investigated the differences between the human annotators and observed that they are caused either by legitimate annotation errors or by the fact that there are multiple valid rationales for a given relation. For example, in the sentence {\em OBJ-PER is the CEO and president of SUBJ-ORG}, the relation \texttt{org:top\_members\/employees} can be explained either by the tokens {\em CEO} or {\em president}. 

The two tables indicate that our approach generates explanations that have considerably higher overlap with human-generated explanations, even though all data points part of this evaluation were chosen to {\em not} have a matching rule. This suggests that our approach generates high-quality explanations of its predictions regardless of whether it has seen the underlying pattern or not. Moreover, the recall of our approach is much higher than that of the other post-hoc explanations, which have not been exposed to rules during training. This shows that with a small amount of supervision, the generated explanations can be better aligned with human intuitions.
The fact that our method outperforms considerably the unsupervised rationale approach of \mbox{\citet{lei-etal-2016-rationalizing}}, which is driven solely by relation classification performance,  further emphasizes that a ``human-in-the-loop'' method such as ours is necessary to yield meaningful explanations.



\begin{figure}[!p]
    \centering
    \footnotesize
    
     \begin{tabularx}{\textwidth}{rX}
     \toprule
      Our Approach  &  The proportion stood at 38.7 percent , down 0.5 percentage points from the first half , said Xia Nong , deputy head of the \hlc[cyan!50]{Industrial Policy Department} \hlc[red]{\textbf{of the}} \hlc[orange!50]{National Development and Reform Commission} ( NDRC ) , at a press conference in Beijing .\\
     &  Gold label: {\tt org:subsidiary}; predicted label: {\tt org:subsidiary} \\
    \midrule
      Saliency & The proportion stood at 38.7 percent , down 0.5 percentage points from the first half , said \hlc[red]{\textbf{Xia Nong}} , deputy head of the \hlc[cyan!50]{Industrial Policy Department} of the \hlc[orange!50]{National Development and Reform Commission} ( NDRC ) , at a press conference in \hlc[red]{\textbf{Beijing}} .\\
      &  Predicted label: {\tt org:subsidiary} \\

      \midrule
      LIME & The proportion stood at 38.7 percent , \hlc[red]{\textbf{down}} 0.5 \hlc[red]{\textbf{percentage}} points from the \hlc[red]{\textbf{first}} half , said Xia Nong , deputy head of the \hlc[cyan!50]{Industrial Policy Department} of the \hlc[orange!50]{National Development and Reform Commission} ( NDRC ) , at a press conference in Beijing .\\
      &  Predicted label: {\tt org:subsidiary} \\
      
      \midrule
      SHAP & The proportion stood at 38.7 \hlc[red]{\textbf{percent , down}} 0.5 percentage points from the first half , said Xia Nong , deputy head of the \hlc[cyan!50]{Industrial Policy Department} of the \hlc[orange!50]{National Development and Reform Commission} ( NDRC ) , at a press conference in Beijing .\\
      &  Predicted label: {\tt org:subsidiary} \\
      
      \midrule
      CXPlain & The proportion stood at 38.7 percent , down 0.5 percentage points from the first half , said Xia Nong , deputy head \hlc[red]{\textbf{of}} the \hlc[cyan!50]{Industrial Policy Department} \hlc[red]{\textbf{of}} the \hlc[orange!50]{National Development and Reform Commission} \hlc[red]{\textbf{(}} NDRC ) , at a press conference in Beijing .\\
      &  Predicted label: {\tt org:subsidiary} \\

      \midrule
      Greedy Adding & The proportion stood at 38.7 percent , down 0.5 percentage points from the first half , said Xia Nong , deputy head of the \hlc[cyan!50]{Industrial Policy Department} \hlc[red]{\textbf{of}} the \hlc[orange!50]{National Development and Reform Commission} ( NDRC ) , at a \hlc[red]{\textbf{press conference}} in Beijing .\\
      &  Predicted label: {\tt org:subsidiary} \\
	\midrule
	  Attention Weights &\raisebox{-\totalheight}{\includegraphics[width=0.9\textwidth]{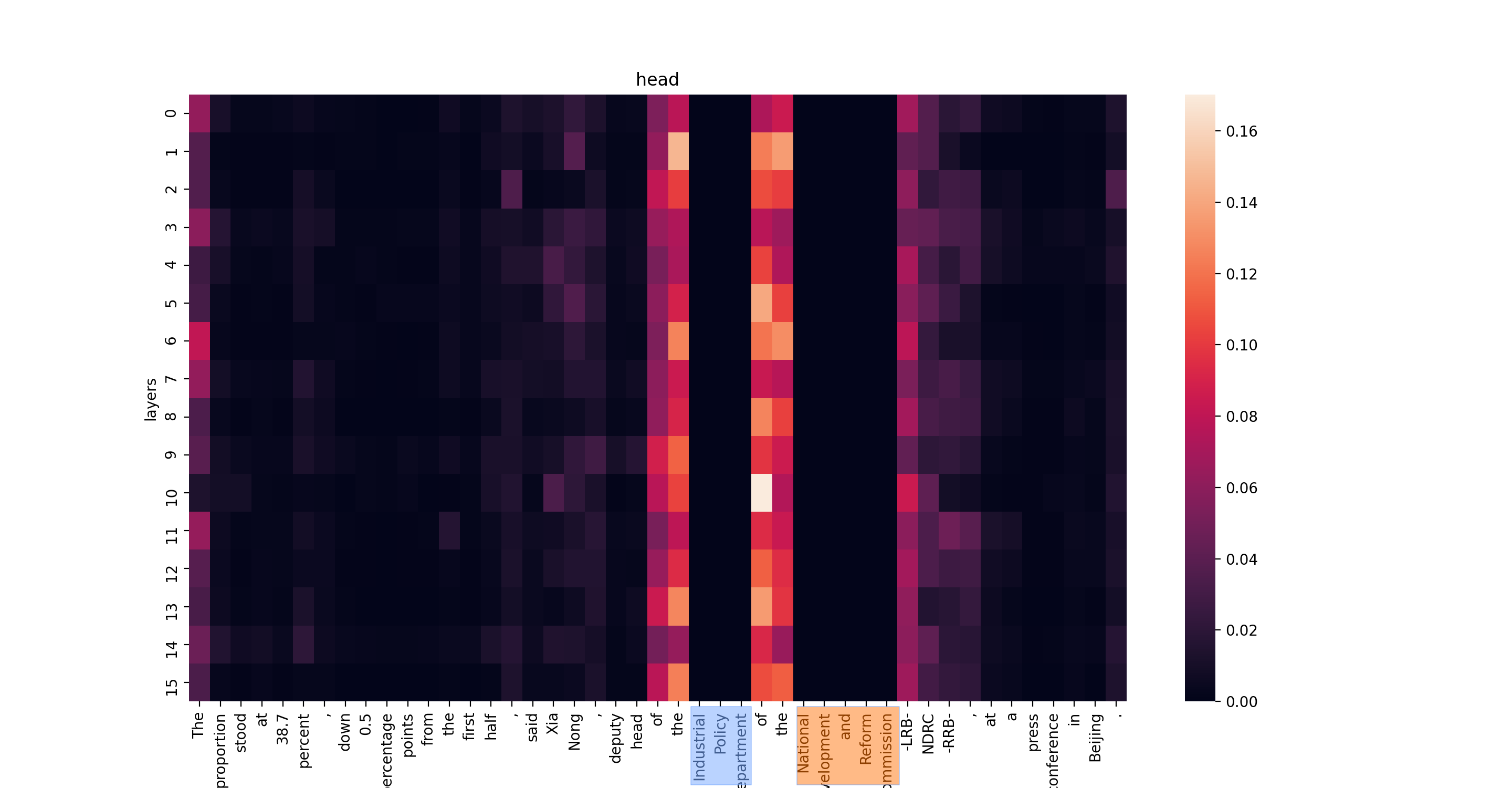}}\\
      \bottomrule
    \end{tabularx}
    \caption{Examples of explainability annotations on TACRED for a {\em correct} RC prediction. The subject and object entities, which are provided in the task input, are highlighted in \textcolor{blue}{blue} and \textcolor{orange}{orange}. The important tokens for explainability identified by the various methods are highlighted in \textcolor{red}{red}. 
    The bottom of the figure shows the heatmap of $[CLS]$ attention weights for BERT's 16 heads (the lighter the color the higher the weight). We shrink the weight range margin to make the color scale distinguishable. For a fair comparison, we masked the subject and object in the attention weights.
   }
    \label{fig:tacred_pos_examples}
\end{figure}

\begin{figure}[!ht]
    \centering
    \footnotesize
    
     \begin{tabularx}{\textwidth}{rX}
\toprule
      Our Approach  &  \hlc[cyan!50]{Pauliina Miettinen} , \hlc[red]{\textbf{a}} defender and goalkeeper \hlc[red]{\textbf{for}} \hlc[orange!50]{Finaland} , was named Tuesday as the new coach of Women 's Professional Soccer inaugural champion Sky Blue .\\
     &  Gold label: {\tt per:origin}; predicted label: {\tt per:countries\_of\_residence} \\
    \midrule
      Saliency & Gonzalez is the \hlc[cyan!50]{Pauliina Miettinen} , a \hlc[red]{\textbf{defender}} and \hlc[red]{\textbf{goalkeeper}} for \hlc[orange!50]{Finaland} , was named Tuesday as the new coach of Women 's Professional \hlc[red]{\textbf{Soccer}} inaugural champion Sky Blue .\\
      &  Predicted label: {\tt per:countries\_of\_residence} \\

      \midrule
      LIME & \hlc[cyan!50]{Pauliina Miettinen} , \hlc[red]{\textbf{a}} defender and goalkeeper for \hlc[orange!50]{Finaland} , was named \hlc[red]{\textbf{Tuesday}} as the new coach of \hlc[red]{\textbf{Women}} 's Professional Soccer inaugural champion Sky Blue .\\
      &  Predicted label: {\tt per:countries\_of\_residence} \\
      
      \midrule
      SHAP & \hlc[cyan!50]{Pauliina Miettinen} , a defender and goalkeeper for \hlc[orange!50]{Finaland} , was \hlc[red]{\textbf{named}} Tuesday \hlc[red]{\textbf{as the}} new coach of Women 's Professional Soccer inaugural champion Sky Blue .\\
      &  Predicted label: {\tt per:countries\_of\_residence} \\
      
      \midrule
      CXPlain & \hlc[cyan!50]{Pauliina Miettinen} \hlc[red]{\textbf{,}} a defender \hlc[red]{\textbf{and goalkeeper}} for \hlc[orange!50]{Finaland} , was named Tuesday as the new coach of Women 's Professional Soccer inaugural champion Sky Blue .\\
      &  Predicted label: {\tt per:countries\_of\_residence} \\

      \midrule
      Greedy Adding & \hlc[cyan!50]{Pauliina Miettinen} , a \hlc[red]{\textbf{defender}} and goalkeeper for \hlc[orange!50]{Finaland} , was named Tuesday as the new coach of Women 's Professional Soccer inaugural \hlc[red]{\textbf{champion}} Sky Blue .\\
      &  Predicted label: {\tt per:countries\_of\_residence} \\
		\midrule
	  Attention Weights &\raisebox{-\totalheight}{\includegraphics[width=0.9\textwidth]{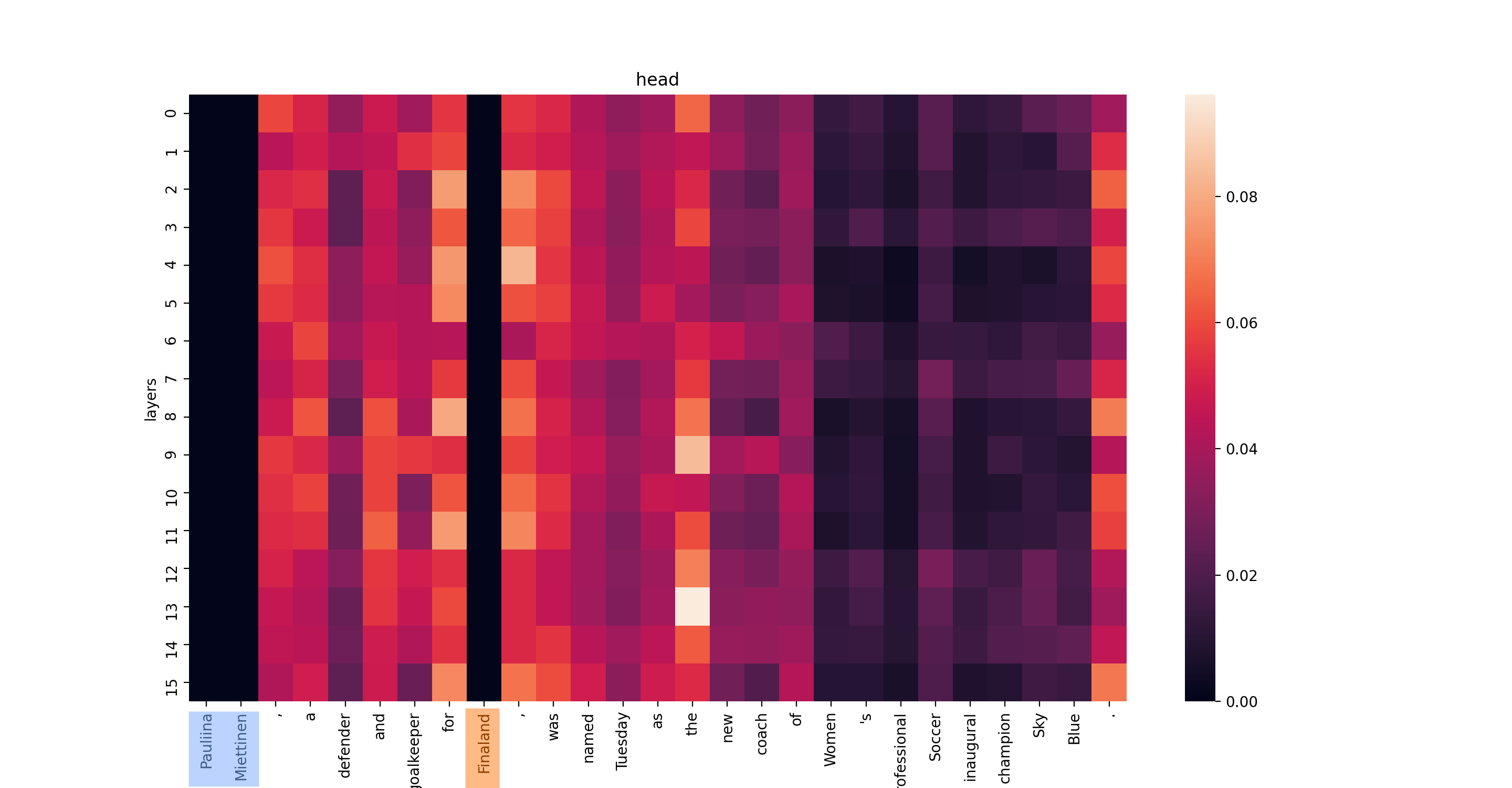}}\\
      \bottomrule
\end{tabularx}
    \caption{ Examples of explainability annotations on TACRED for an {\em incorrect} RC prediction. This figure follows the same convention as Figure~\ref{fig:tacred_pos_examples}.
   }
    \label{fig:tacred_neg_examples}
\end{figure}

\begin{figure}[!ht]
    \centering
    \footnotesize
    
     \begin{tabularx}{\textwidth}{rX}
     \toprule
      Our Approach  &` `  The U.S. has the biggest stock of chemical arms in the world , and it is trying to obstruct other countries from having their own , ' ' said \hlc[orange!50]{Arab League} \hlc[red]{\textbf{Secretary-General}} \hlc[cyan!50]{Chedli Klibi} .\\
     &  Gold label: {\tt Work\_For}; predicted label: {\tt Work\_For} \\
    \midrule
      Saliency &\hlc[red]{\textbf{`}} ` The U.S. has the biggest stock of chemical arms in the world , and it is trying to obstruct other countries from having their own , ' ' said \hlc[orange!50]{Arab League} Secretary-General \hlc[cyan!50]{Chedli Klibi} .\\
      &  Predicted label: {\tt Work\_For} \\

      \midrule
      LIME &` ` The U.S. has the biggest stock of chemical arms in the world , and it is trying to obstruct other countries from having their own \hlc[red]{\textbf{,}} ' ' said \hlc[orange!50]{Arab League} Secretary-General \hlc[cyan!50]{Chedli Klibi} .\\
      &  Predicted label: {\tt Work\_For} \\
      
      \midrule
      SHAP &` ` The U.S. has the biggest stock of chemical arms in the world , and it is trying to obstruct other countries from having their own , ' ' \hlc[red]{\textbf{said}} \hlc[orange!50]{Arab League} Secretary-General \hlc[cyan!50]{Chedli Klibi} .\\
      &  Predicted label: {\tt Work\_For} \\
      
      \midrule
      CXPlain &` ` The U.S. has the biggest stock of chemical arms in the world , and it is trying to obstruct other countries from having their own , ' \hlc[red]{\textbf{'}} said \hlc[orange!50]{Arab League} Secretary-General \hlc[cyan!50]{Chedli Klibi} .\\
      &  Predicted label: {\tt Work\_For} \\

      \midrule
      Greedy Adding &` ` The U.S. has the biggest stock of chemical arms in the world , and it is trying to obstruct other countries from having their own , ' ' said \hlc[orange!50]{Arab League} \hlc[red]{\textbf{Secretary-General}} \hlc[cyan!50]{Chedli Klibi} .\\
      &  Predicted label: {\tt Work\_For} \\
       \midrule
	 Attention Weights &\raisebox{-\totalheight}{\includegraphics[width=0.9\textwidth]{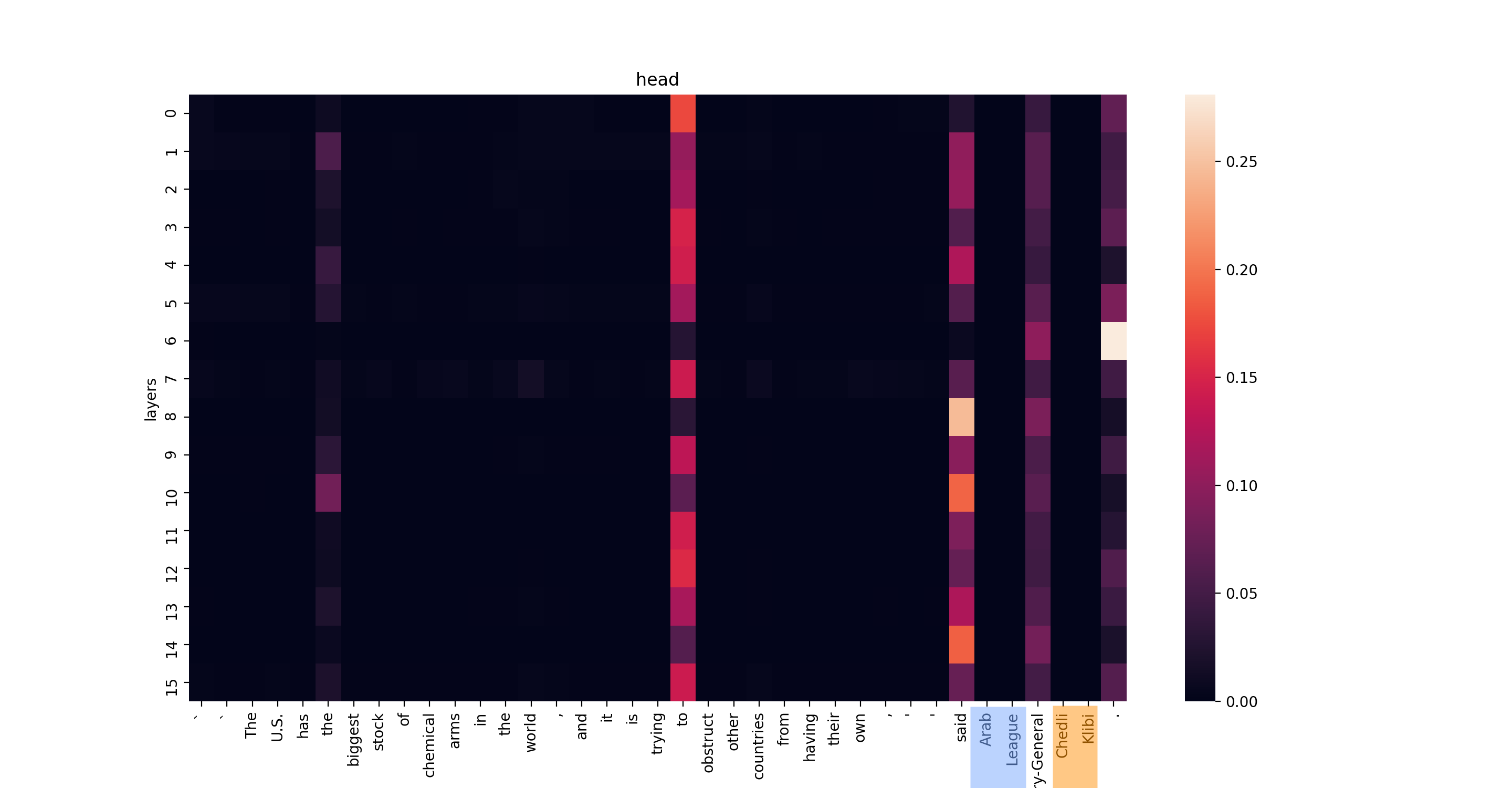}}\\
      \bottomrule
    \end{tabularx}
    \caption{Examples of explainability annotations on CoNLL04 for a {\em correct} RC prediction. This figure follows the same convention as Figure~\ref{fig:tacred_pos_examples}.
   }
    \label{fig:conll_pos_examples}
\end{figure}

\begin{figure}[!ht]
    \centering
    \footnotesize
    
     \begin{tabularx}{\textwidth}{rX}
      \toprule
      Our Approach  &  ` ` We 're in for a long haul , ' ' said \hlc[cyan!50]{Dave Olson} \hlc[red]{\textbf{of}} the \hlc[orange!50]{Payette National Forest} in Idaho , where more than 200 fires continued to burn.\\
     &  Gold label: {\tt no\_relation}; predicted label: {\tt Work\_For} \\
    \midrule
      Saliency & ` ` We 're in for a long haul , ' ' said \hlc[cyan!50]{Dave Olson} of the \hlc[orange!50]{Payette National Forest} in \hlc[red]{\textbf{Idaho}} , where more than 200 fires continued to burn.\\
      &  Predicted label: {\tt Work\_For} \\

      \midrule
      LIME & ` ` We 're in for a long haul , ' ' \hlc[red]{\textbf{said}} \hlc[cyan!50]{Dave Olson} of the \hlc[orange!50]{Payette National Forest} in Idaho , where more than 200 fires continued to burn.\\
      &  Predicted label: {\tt Work\_For} \\
      
      \midrule
      SHAP & ` ` We \hlc[red]{\textbf{'re}} in for a long haul , ' ' said \hlc[cyan!50]{Dave Olson} of the \hlc[orange!50]{Payette National Forest} in Idaho , where more than 200 fires continued to burn.\\
      &  Predicted label: {\tt Work\_For} \\
      
      \midrule
      CXPlain & ` ` We 're in for a long haul , ' ' said \hlc[cyan!50]{Dave Olson} of the \hlc[orange!50]{Payette National Forest} in \hlc[red]{\textbf{Idaho}} , where more than 200 fires continued to burn.\\
      &  Predicted label: {\tt Work\_For} \\

      \midrule
      Greedy Adding & ` ` We 're in for a long haul , \hlc[red]{\textbf{' ' said}} \hlc[cyan!50]{Dave Olson} of the \hlc[orange!50]{Payette National Forest} \hlc[red]{\textbf{in}} Idaho , where more than \hlc[red]{\textbf{200}} fires \hlc[red]{\textbf{continued}} to burn.\\
      &  Predicted label: {\tt Work\_For} \\
	\midrule
	  Attention Weights &\raisebox{-\totalheight}{\includegraphics[width=0.9\textwidth]{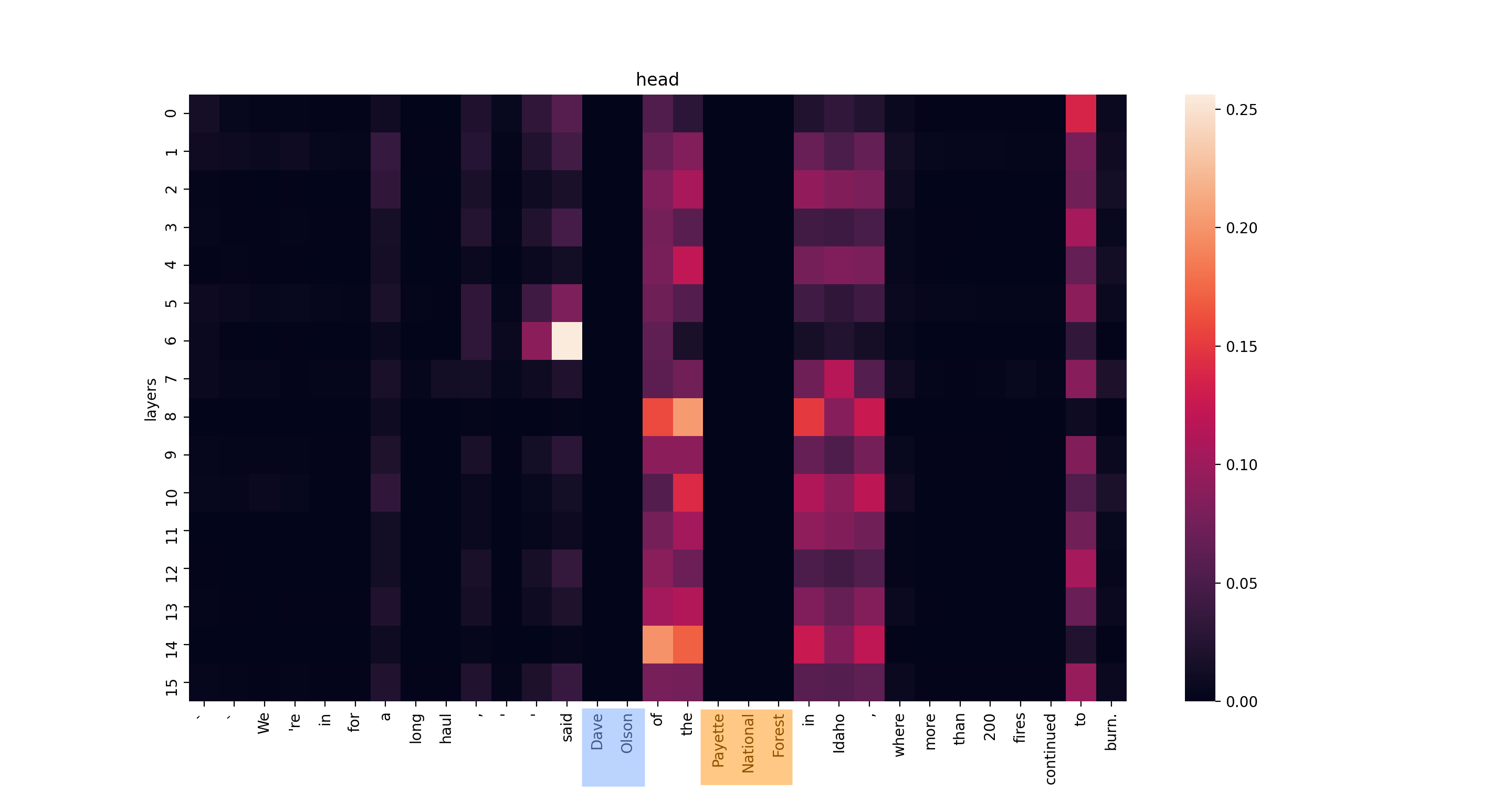}}\\
      \bottomrule
    \end{tabularx}
    \caption{Examples of explainability annotations on CoNLL04 for an {\em incorrect} RC prediction. This figure follows the same convention as Figure~\ref{fig:tacred_pos_examples}.
   }
    \label{fig:conll_neg_examples}
\end{figure}

We include several examples of the generated rationales in Figures~\ref{fig:tacred_pos_examples},~\ref{fig:tacred_neg_examples},~\ref{fig:conll_pos_examples}, and~\ref{fig:conll_neg_examples}.  These examples indicate that most of the baselines are noisier, i.e., they contain a considerable amount of false positives (words that should not be part of the rationale) and false negatives (words that should be included but are not). In contrast, our method does a better job focusing on the right explanation tokens. 

In the example in Figure~\ref{fig:tacred_pos_examples}, both our RC model and vanilla BERT predicted the correct relation. However, our method labels only the preposition {\em of} and the determiner {\em the} as its explanation, while other baselines such as LIME and SHAP completely missed them. Greedy adding and CXPlain label more irrelevant words in the context such as {\em (} and {\em press conference}. The attention weights do capture the key words, but we can clearly see additional noise surrounding the entities.
In the example in Figure~\ref{fig:tacred_neg_examples}, both our model and vanilla model predicted the incorrect relation. Our model labels the preposition {\em for}, which provides a strong hint  for its (possibly) incorrect prediction ({\tt per:countries\_of\_residence}). In contrast, the baselines focus more on the nouns such as {\em defender} and {\em champion}. Applying the substitution heuristic indicates that the preposition {\em for} is necessary for the explanation (e.g., changing it to {\em against} changes the relation), while the nouns are not relevant. In this example, the attention weights are almost completely noisy.

In Figure~\ref{fig:conll_pos_examples}, both our model and the vanilla SpanBERT model produce the correct prediction. The words {\em Secretary-General} clearly explain the {\em Work\_For} relation in the explanations generated by our model and greedy adding. The other baselines do not provide meaningful explanations here.
In Figure~\ref{fig:conll_neg_examples}, which shows an incorrect prediction, only our model can defend its prediction by its explanation. The baseline approaches cannot provide valid explanations to defend the prediction at all. We also find that with the explanation provided from our model, one can argue that the predicted relation is actually correct, and we should change the gold label instead.

\begin{figure}[ht!]

\begin{tabular}{c}
\includegraphics[width=0.75\textwidth]{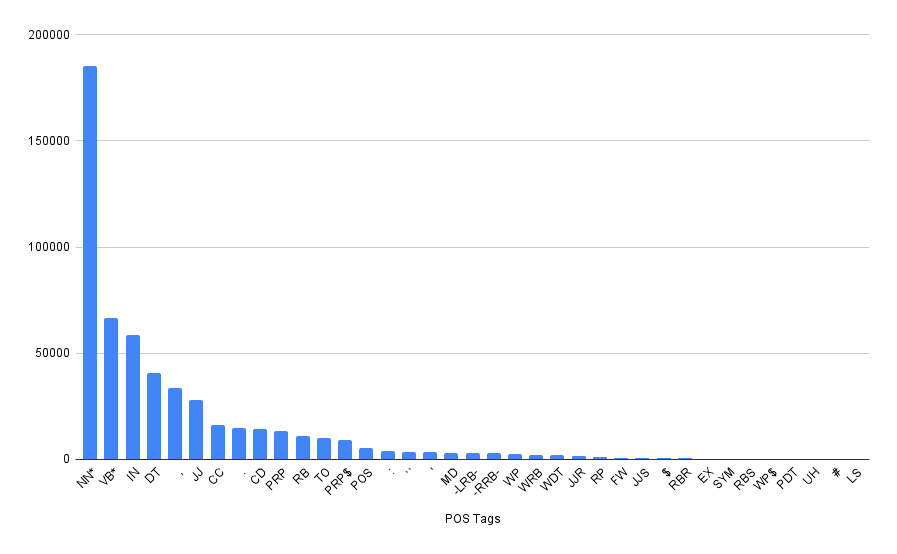}\\
Counts of POS tags in the test partition.\\
\includegraphics[width=0.75\textwidth]{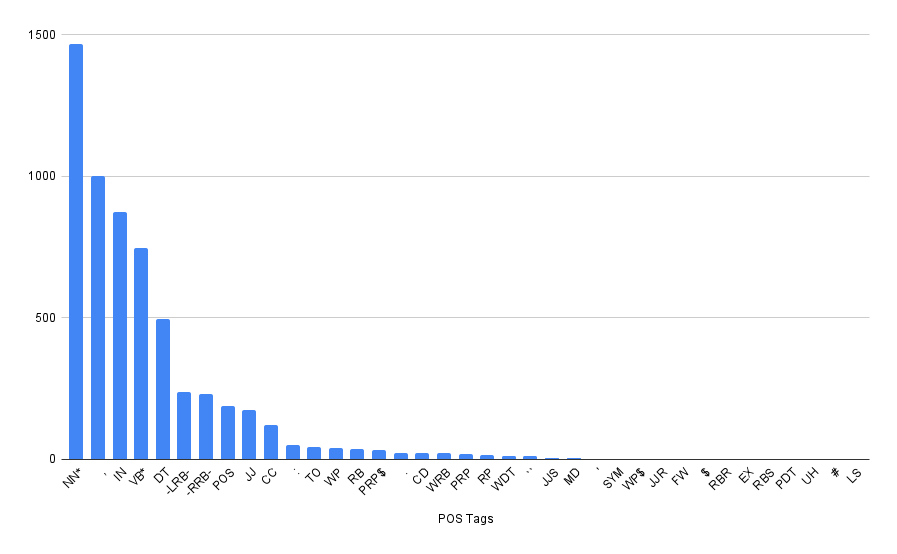}\\
Counts of POS tags in the  explanations in the test partition.\\
\end{tabular}
\vspace{2mm}

  
%
\caption{The distributions of POS tags in the TACRED test partition. The top figure shows how many times each POS tag appears in the test data. The bottom figure shows how many times each POS tag appears in the generated explanations from the same partition.
}
\label{fig:postags2}
\end{figure}

\citet{lei-etal-2016-rationalizing} state that rationales should be short, coherent, and be sufficient for the correct prediction. However, short does not necessarily mean simple. To highlight this point, Figure~\ref{fig:postags2} compares the distribution of POS tags in the TACRED test partition with the distribution of POS tags that participate in explanations in the same partition. We draw two observations from this data. First, to extract plausible rationales, our EC has to diverge from the distribution of POS tags in the data in a non-trivial way. For example, the frequency of verbs (\texttt{VB*}), prepositions (\texttt{IN}), and commas is considerably higher in the explanations than the raw data. 
Second, the figure indicates that our explanations often focus on parts of speech that are necessary for plausability (according to the human annotators) but are semantically-ambiguous such as prepositions (\texttt{IN}), commas,\footnote{Commas are necessary to capture appositive constructs, which are often indicative of relations, e.g., ``Barack Obama, the former president.'' In cases such as these, the subject and object of the relation (e.g., ``Barack Obama'' and ``former president'', respectively) cover most lexical information relevant to the relation. In these cases, the remaining signal that indicates the apposition is the comma.} and determiners (\texttt{DT}). This is different from traditional pattern acquisition methods \cite{riloff1996automatically}, which usually focus on words with more clear semantics such as nominals and verbs.\footnote{Note that traditional patterns may include prepositions and particles, e.g., in verb constructs such as \texttt{SUBJECT was born in OBJECT}. However, these patterns are usually semantically headed by verb phrases or nominalized predicates, e.g., \texttt{born}, and seldom by prepositions.}

\subsubsection{Ablation Study}

\begin{table*}[]
\centering
\begin{tabular}{lccc}
& RC F1 & Quantitative EC F1 & Qualitative EC F1 \\ 
\hline
Full Model &  {\bf 70.52{\small$\pm 0.54$}}  &  {\bf 95.76} &   {\bf 62.05}    \\
\qquad$-$ NRC & 67.47 {\small$\pm 0.54$} & 92.95 & 54.70  \\
\qquad$-$ EC & 70.62{\small$\pm 0.46$}    & N/A & N/A\\
Vanilla SpanBERT & 70.07{\small$\pm 0.73$}    & N/A & N/A\\
\end{tabular}
\caption{Ablation results on the TACRED test partition, i.e., ``--'' indicates that the corresponding component was removed from the full system, and ``N/A'' indicates that that metric is not applicable.}
\label{ablation_tac}
\end{table*}

\begin{table*}[]
\centering
\begin{tabular}{lccc}
& RC F1 & Quantitative EC F1 & Qualitative EC F1 \\  
\hline
Full Model &  {\bf 79.46{\small$\pm 0.92$}}  & {\bf 99.52}  & {\bf 58.97}\\
\qquad$-$ NRC & 77.34{\small$\pm 2.33$} & 99.00 & 50.12 \\
\qquad$-$ EC & 76.58{\small$\pm 1.52$}    & N/A & N/A\\
Vanilla SpanBERT & 75.78{\small$\pm 4.79$} & N/A & N/A\\
\end{tabular}
\caption{Ablation results on the CoNLL04 test partition, i.e., ``--'' indicates that the corresponding component was removed from the full system, and ``N/A'' indicates that that metric is not applicable.}
\label{ablation_conll04}
\end{table*}

To understand the impact of the classifiers employed by our approach (i.e., NRC, RC, and EC), we implemented ablation experiments on both datasets, which are summarized in Tables~\ref{ablation_tac} and~\ref{ablation_conll04}. 
Note that the method without both NRC and EC becomes equivalent to the vanilla SpanBERT
(as we discussed in Section~\ref{sec:baselines_rc}).
Overall, this experiment re-emphasizes that not only does our approach outperform the vanilla SpanBERT, but it does so while generating an explanation for its decisions.
Removing the NRC drops the relation classification F1 score by approximately 3 points on TACRED, and 2 points on CoNLL04. This impact is explained by that fact the using the NRC avoids the meaningless scenario where the EC (which was trained only on positive examples) is applied to negative examples. 
Interestingly, removing the EC has no statistical impact on relation classification performance on TACRED, but it reduces the relation classification F1 by approximately 3 points on CoNLL04. As discussed in Section~\ref{sec:rc}, this is caused by the fact that the EC serves as a useful disambiguator in CoNLL04, where multiple relations co-occur in the same sentence. The EC is not that impactful in TACRED, which has a more artificial setting with much fewer relations per sentence.\footnote{The average number of relations per sentence in TACRED is approximately 2 in training, and 1 in development and test.}

\subsubsection{Interpretability: from Local to Global}

\begin{table*}[]
\centering
\begin{tabular}{rccc}
Approach            & Precision & Recall & F1    \\ \hline
\multicolumn{4}{c}{Baseline} \\
\hline
Manual Rules\textsuperscript{[1]} & {\bf 85.93} & 24.24         & 37.81       
\\
\hline
\multicolumn{4}{c}{Our Approach} \\
\hline
Rules from Training\textsuperscript{[2]} &  49.39  &  30.26 &   37.52    \\
Rules from Test\textsuperscript{[3]} & 59.69     & 55.04 & 57.27 \\
Combination of [1] and [2] & 54.12 & 62.95 & 58.20\\
Combination of [1] and [3] & 65.28 & 71.64 & {\bf 68.31}\\
Combination of [2] and [3] & 56.34 & 40.90 & 47.40\\
Combination of [1], [2] and [3] & 57.36     &{\bf 72.00} &  63.85 \\
\end{tabular}
\caption{ Performance of the rule-based model on the TACRED test partition. [1] is the set of manually-written surface rules of~\citet{Angeli2015BootstrappedST} coupled with our syntactic rules (see Section~\ref{sec:dataprep}). [2] is the set of rules generated from our explainability classifier's outputs with gold labels on the training partition. [3] is the set of rules from the explainability classifier's outputs with predicted labels on the test partition. We also evaluate the performance on combinations of these sets of rules: [2]+[3] contain all rules generated by our approach; [1]+[2]+[3] combine machine-generated rules with the manually-written rules.
}
\label{results_rule_tac}
\end{table*}

\begin{table*}[]
\centering
\begin{tabular}{rccc}
Approach            & Precision & Recall & F1    \\ \hline
\multicolumn{4}{c}{Baseline} \\
\hline
Manual Rules\textsuperscript{[1]} &   {\bf 81.82} & 17.06         & 28.24       
\\
\hline
\multicolumn{4}{c}{Our Approach} \\
\hline
Rules from Training\textsuperscript{[2]} &   66.10 & 27.73 & 39.07\\
Rules from Test\textsuperscript{[3]} & 67.95 &50.24 & 57.77 \\
Combination of [1] and [2] & 71.06     & 39.57 & 50.84\\
Combination of [1] and [3] & 68.48 & 59.72 & 63.80\\
Combination of [2] and [3] & 64.01 & 55.21 & 59.29 \\
Combination of [1], [2] and [3] & 66.67     & {\bf 63.03}&  {\bf 64.80} \\
\end{tabular}
\caption{ Performance of the rule-based model on the CoNLL04 test partition. This table follows the same conventions as Table~\ref{results_rule_tac}, except, in this case, [1] is the set of manually-written Odin rules we wrote for CoNLL04. 
}
\label{results_rule_col}
\end{table*}

Lastly, we evaluate the performance of our rule-based model that relies solely on rules, some of which were manually written (see Section~\ref{sec:dataprep}), while some were automatically generated by our approach, as described in Section~\ref{sec:global}. The results are summarized in Tables~\ref{results_rule_tac} and~\ref{results_rule_col}.
We draw two observations from these results:

\begin{itemize}
\item Automatically-generated rules can outperform manually-written ones. However, in order to approach the performance of the neural RC, our method benefits from being aware of the distribution of words in each testing sentence to be processed (setting [3] in the tables). Importantly, we reiterate that when using the test sentences, our approach does {\em not} have access to any gold human annotations for RC and EC. That is, the rules generated from test sentences rely only on predicted relation labels and predicted explanations for each given sentence.
The fact that rules need to be exposed to more data before they generalize is not extremely surprising: the rule matching engine we currently use relies on exact lexical matching, which means that the actual tokens to be matched must be present in the rule. However, the fact that the knowledge necessary to encode a relation extraction {\em can} be encoded into rules is exciting. The combination of these observations suggests that a future avenue for research that focuses on ``soft rule matching''~\cite{zhou2020nero}, might be the direction that captures the advantages of both rules and neural methods. 

\item Interestingly, automatically-generated rules tend to be complementary to the manual ones. The combination of all three rule sets ([1], [2], and [3] in the tables) outperforms considerably both the setting that relies solely on manual rules and the configuration that relies only on automatically-generated ones. The combination of all rule sets outperforms the manually-generated rules by 31\% F1 and 38\% F1 (absolute) in TACRED and CoNLL04, respectively. Furthermore, the TACRED result of the combined rule set approaches the performance of the neural RC within less than 3\% F1. The performance gap between the combined rule set and neural RC in CoNLL04 is larger (over 14\% F1).\footnote{We conjecture that the cause for this larger gap is the lower quality of the rules used for the CoNLL04 dataset. That is, the TACRED rules were developed by a larger team over a longer period of time, whereas the CoNLL04 rules were developed by one of the authors in only a few hours.} Nevertheless, all in all, this result suggests that humans and machines can collaborate towards building a fully-explainable model that comes reasonably close to the performance of neural classifiers. 

\end{itemize}



\subsubsection{Error Analysis}
\label{error_analysis}

\begin{table}[t]
    \centering
    \footnotesize
    
     \begin{tabularx}{\textwidth}{X}
      \toprule
       \hlc[orange!50]{Rio de Janeiro} \hlc[cyan!50]{O GLOBO}\\
      Gold label: {\tt OrgBased\_In}; predicted label: {\tt OrgBased\_In} \vspace{3mm} \\      
    \midrule
     I had an e-mail exchange with \hlc[cyan!50]{Benjamin Chertoff} of Popular Mechanics in the original Loose Change thread that showed that he was not a close relative of \hlc[orange!50]{Michael Chertoff} .\\
      Gold label: {\tt no\_relation}; predicted label: {\tt per:other\_family} \vspace{3mm} \\
      \midrule
      In Beijing Thursday, \hlc[red]{\textbf{spokesman}} \hlc[cyan!50]{Li} repeated \hlc[orange!50]{China}'s position that the key to the solution of the Cambodian conflict `` lies in the genuine and complete Vietnamese troop withdrawal at the earliest possible date and effective international supervision. ''\\
      Gold label: {\tt Live\_in}; predicted label: {\tt Work\_for} \vspace{3mm} \\
      \midrule
      There's been a sea change in my lifetime, `` said \hlc[orange!50]{Jefferson Keel} , lieutenant governor of the Chickasaw Nation in Oklahoma and a first vice \hlc[red]{\textbf{president}} of the \hlc[cyan!50]{National Congress of American Indians}. ''\\
      Gold label: {\tt no\_relation}; predicted label: {\tt org:top\_members/employees} \vspace{3mm} \\
      \midrule
     Overview of Arrow Missile Program , Tasks \hlc[red]{\textbf{94AA0008Z}} \hlc[orange!50]{Jerusalem} \hlc[cyan!50]{THE JERUSALEM POST} in English 15 Oct 93 pp 6-8 , 10--FOR OFFICIAL USE ONLY\\
      Gold label: {\tt OrgBased\_In}; predicted label: {\tt OrgBased\_In}  \vspace{3mm} \\
      \midrule
     Like al-Shabab , \hlc[red]{\textbf{the}} \hlc[cyan!50]{ADF} \hlc[red]{\textbf{is primarily a}} \hlc[orange!50]{Muslim} \hlc[red]{\textbf{radical}} group .\\
     Gold label: {\tt org:political\/religious\_affiliation};\\
     predicted label: {\tt org:political\/religious\_affiliation}  \vspace{3mm} \\
     \bottomrule
    \end{tabularx}
    \caption{Typical errors that our explainability classifier commits. These include errors of under prediction (first two rows), misleading prediction (middle two rows), and errors of over prediction (last two rows). This figure follows the same convention as Figure~\ref{fig:tacred_pos_examples}.}
    \label{tab:errors}
\end{table}

We conclude this section with a brief error analysis of our explainability classifier in the TACRED and CoNLL04 datasets. Table~\ref{tab:errors} summarizes a few typical errors observed in the two datasets. 

The first two rows in the table show examples where the EC generates explanations that rely solely on the subject and object entities, without including any word in the relations' contexts. Note that the example shown in the first row is potentially correct: it is likely that a location name that immediately precedes an organization name indicates the location of that organization. However, the second example is clearly incorrect: the correct explanation to justify the {\tt no\_relation} label should minimally include {\em not} and {\em relative}. 
Further, please note that a hypothetical RC that had access to the {\em unmasked} entities could potentially perform even better. For example, in the first case, one could infer that {\em O Globo} is based in {\em Rio de Janeiro} because the former organization name is Portuguese. However, our RC only sees masked subjects and objects. Nevertheless, we believe that our strategy of masking entities participating in relations is a valuable exercise, as it investigates the capacity of neural methods to identify explicit context necessary for relation extraction. 

Rows 3 and 4 in the table show examples when our RC makes incorrect predictions due to incorrect tokens labeled by the EC.
For example, the token {\em president} in row 4 guides the RC towards the incorrect prediction {\tt org:top\_members/employees}.
The situation in the third row is more subtle: one might argue that {\em China} here can also be referring to the government, which makes the prediction {\tt Work\_for} correct. 
In any case, these errors indicate that our explanations can be used for debugging purposes when the RC makes incorrect predictions.

The last two rows in the table show examples where our EC over included words in its explanations. For example, in the last row, a likely interpretation is that the verb {\em is}  should be part of the correct explanation, but all the other words are unnecessary. This happens because the rule lexical triggers in TACRED tend to contain multiple words, which encouraged the EC to learn to include additional words in its explanation. In contrast, in CoNLL04 (second to last row), most triggers are single-word phrases. This prompted the EC to include one token in its explanation, even though it is unnecessary for the prediction of the relation label in this case. 

For a more complete bigger picture, we analyzed the overall frequency of these error types on the same sampled instances we used for the qualitative explanation evaluation (Section~\ref{sec:qualeval}).
Errors where the EC provided no explanations\footnote{We included in this category the situations where the explanation was completely empty or it included only the subject and/or object entity mentions.} occurred in 4.12\% of examples in TACRED, and 19.41\% in CoNLL04. Errors where the explanations caused false positive relations to be predicted appeared 25.95\% times in TACRED, and 16.49\% in CoNLL04. 
Nevertheless, as Tables~\ref{results_2_tac},~\ref{results_2_conll},~\ref{results_human_tacred}, and~\ref{results_human_conll} show, our EC makes considerably fewer errors than all other explainability methods. There is no reason to believe that its current errors cannot be fixed with human feedback that would provide a (hopefully small) number of rules to adjust imperfect explanations.


%% file: sections/conclusion.tex
\section{Conclusion}
\label{sec:conclusion}
We introduced an explainable approach for relation extraction that jointly trains for prediction and explainability. 
Our approach uses a multi-task learning framework with a shared encoder, and jointly trains a classifier for relation extraction with a second explainability classifier that labels which words in the context of the relation explain the underlying relation. Further, our method is semi-supervised, as annotations for the latter classifier are usually not available. 

We evaluated the proposed approach on a relation extraction task in two datasets: TACRED and CoNLL04. Our evaluation showed that, even with minimal supervision for explanation guidance, our method generates explanations for the relation classifier's decisions that are considerably more accurate and plausible than other strong baselines such as LIME, or relying on attention weights \cite{simonyan2013deep,bahdanau2014neural,ribeiro2016should,NIPS2017_7062, schwab2019cxplain,vafa2021rationales}. 
Further, our results indicated that jointly training for explainability and prediction improves the prediction task itself, i.e., the relation classifier performs better when it is exposed only to the textual context deemed important by the explainability classifier.

We also showed that it is possible to convert these local explanations into global ones. We converted the outputs of our explainability classifier into a set of rules that globally explains the behavior of the neural relation classifier. 
Our results showed that our strategy for generating a rule-based model pushes the performance of rule-based approaches closer to that of neural methods.


Longer term, we envision our approach being used in an iterative semi-supervised learning scenario akin to co-training~\cite{blum1998combining}. That is, the newly generated rules can be converted to executable rules that can be applied over large, unannotated texts to generate new training examples for the relation classifier, and vice versa.
Further, our method could potentially benefit from traditional pattern bootstrapping approaches~\cite{riloff1996automatically,Lin2001DIRTD}, which could reduce the amount of human supervision necessary by automatically expanding the set of initial patterns available.

At a higher level, we hope that this work will support meaningful collaborations between NLP researchers and subject matter experts in other domains (e.g., medical, legal), who benefit from the output of NLP systems (e.g., large-scale extraction of biomedical events) but may not understand the intricacies of the neural methods that underlie these NLP approaches. 

We release all code and data behind this work at: \url{https://github.com/clulab/releases/cl2022-twoflints/}

%% file: sections/appendix.tex
\appendixsection{Experimental Details}
\label{appendix}
We use the dependency parse trees, POS tags and NER labels as included in the original release of the TACRED dataset. All these were generated with Stanford CoreNLP~\cite{manning-EtAl:2014:P14-5}.

We use the pretrained SpanBERT model ~\cite{joshi-etal-2020-spanbert} available in the HuggingFace transformer library~\cite{wolf-etal-2020-transformers} as our encoder.\footnote{\url{https://huggingface.co/SpanBERT/spanbert-large-cased}} Table~\ref{modeldetails} shows the hyperparameter details for training the neural models for relation classification (SpanBERT) and both relation and explainability classification (Unsupervised Rationale and our approach). Note that we relied mostly on the default hyperparameter values from SpanBERT, but used a larger number of epochs with a smaller learning rate to fine-tune the additional explainability component.
 The Unsupervised Rationale method was tuned for relation classification, which boosted its RC performance (Tables~2 and~3), but negatively impacted its explainability power (Tables~4 and~5).

\begin{table*}[]
\centering
\begin{tabular}{rccc}
Approach & SpanBERT & Unsupervised Rationale & Our Approach \\
\hline
Number of epochs & 10$^\ast$ & 20 & 20 \\
Learning rate & 2e-5$^\ast$ & 1e-5 & 1e-5 \\
Dropout rate & 0.1$^\ast$ & 0.1 & 0.1 \\
Batch size & 32$^\ast$ & 32 & 32 \\
Max sequence length & 128$^\ast$ & 128 & 128 \\
Scheduler & \multicolumn{3}{c}{Linear scheduler with warm up$^\ast$} \\ 
\end{tabular}
\caption{Hyperparameter details for training the neural models for relation classification (for SpanBERT) and both components (Unsupervised Rationale and our approach). The numbers with $\ast$ are the default values from the SpanBERT implementation available at: \url{https://github.com/facebookresearch/SpanBERT}.}
\label{modeldetails}
\end{table*}

 Some of the explainability baselines do not have hyper parameters, including: attention, saliency mapping, greedy adding, and all words in between. For SHAP, we use all default settings from the API provided by the authors at: \url{https://shap.readthedocs.io/en/latest/index.html}. For LIME, the number of samples we used is 2000. And for CXPlain, the explanation model we use is a 2-layers RNN model, with learning rate of 0.001, dropout rate of 0.2, and trained for 2 epochs. 